%% file: main.tex
\documentclass[10pt,journal,compsoc]{IEEEtran}

\usepackage[nocompress]{cite}  

\usepackage{xcolor}  
\definecolor{mydarkblue}{rgb}{0.68, 0.85, 1.0}

\usepackage{microtype}
\usepackage{graphicx}
\usepackage{booktabs} 
\usepackage{soul}

\usepackage[colorlinks=true,
    linkcolor=mydarkblue,
    citecolor=mydarkblue,
    filecolor=mydarkblue,
    urlcolor=mydarkblue]{hyperref}
\usepackage{hyperref}

\usepackage{amsmath}
\usepackage{amssymb}
\usepackage{mathtools}
\usepackage{amsthm}
\usepackage{graphicx}

\usepackage[capitalize,noabbrev]{cleveref}

\usepackage{hyperref}
\usepackage{url}

\usepackage{amsmath}
\usepackage{graphicx}
\usepackage{amssymb}
\usepackage{wrapfig}
\usepackage{multirow}
\usepackage{tabularx}
\usepackage{adjustbox}
\usepackage{color, colortbl}
\usepackage{float} 
\usepackage{makecell}
\usepackage{booktabs} 
\usepackage{caption}
\usepackage{subcaption}
\usepackage{amsfonts}
\usepackage{enumerate}
\usepackage{bm}
\usepackage{slashbox}
\usepackage{diagbox}
\usepackage[marginal]{footmisc}
\usepackage{sidecap}
\usepackage[english]{babel}

\theoremstyle{plain}

\theoremstyle{definition}

\theoremstyle{remark}

\usepackage{color, colortbl}
\definecolor{Gray}{gray}{0.9}

\newcommand{\x}{\mathbf{x}}

\newcommand{\y}{\mathbf{y}}
\newcommand{\z}{\mathbf{z}}

\newcommand{\btheta}{\boldsymbol{\theta}}
\newcommand{\bphi}{\boldsymbol{\phi}}
\newcommand{\bpsi}{\boldsymbol{\psi}}

\usepackage{tikz}
\usepackage{pifont}
\definecolor{mColor2}{rgb}{0.95,0.95,0.95}
\newcommand{\etal}{\textit{et al.}}

\ifCLASSINFOpdf
\else
\fi
\begin{document}
%
\title{Beyond Model Adaptation at Test Time: A Survey}
%

%

\author{Zehao~Xiao,
        and Cees~G.~M.~Snoek, ~\IEEEmembership{Senior Member,~IEEE}
\IEEEcompsocitemizethanks{
\IEEEcompsocthanksitem Z. Xiao and C.G.M. Snoek are with the University of Amsterdam, the Netherlands. E-mail: \{z.xiao, c.g.m.snoek\}@uva.nl. \protect
}
\thanks{Manuscript received, 2024; revised, 2024.}
}

\input{0_abstract}
\maketitle

\IEEEdisplaynontitleabstractindextext

\IEEEpeerreviewmaketitle
\input{1_introduction}
\input{2_background}
\input{3_methods_what}
\input{3_methods_how}
\input{4_application}
\input{6_discussion}
\input{7_conclusion}


\bibliography{main}

\ifCLASSOPTIONcaptionsoff
  \newpage
\fi

\end{document}

%% file: 0_abstract.tex
\IEEEtitleabstractindextext{%
\begin{abstract}

Machine learning algorithms have achieved remarkable success across various disciplines, use cases and applications, under the prevailing assumption that training and test samples are drawn from the same distribution. Consequently, these algorithms struggle and become brittle even when samples in the test distribution start to deviate from the ones observed during training. Domain adaptation and domain generalization have been studied extensively as approaches to address distribution shifts across test and train domains, but each has its limitations. Test-time adaptation, a recently emerging learning paradigm, combines the benefits of domain adaptation and domain generalization by training models only on source data and adapting them to target data during test-time inference. In this survey, we provide a comprehensive and systematic review on test-time adaptation, covering more than 400 recent papers. We structure our review by categorizing existing methods into five distinct categories based on what component of the method is adjusted for test-time adaptation: the model, the inference, the normalization, the sample, or the prompt, providing detailed analysis of each. We further discuss the various preparation and adaptation settings for methods within these categories, offering deeper insights into the effective deployment for the evaluation of distribution shifts and their real-world application in understanding images, video and 3D, as well as modalities beyond vision. We close the survey with an outlook on emerging research opportunities for test-time adaptation. 
A list of test-time adaptation literature is provided at 
\url{https://github.com/zzzx1224/Beyond-model-adaptation-at-test-time-Papers}.
%
%
\end{abstract}
}

%% file: 1_introduction.tex
\section{Introduction}\label{sec:introduction}



\IEEEPARstart{M}{achine} learning has achieved a remarkable success in a wide variety of disciplines, use cases and applications. 
For instance, AlexNet \cite{krizhevsky2012imagenet} demonstrated that training deep convolutional networks at scale outperforms traditional vision feature-engineering. LSTM \cite{hochreiter1997long} advanced sequential prediction by effectively handling long-term dependencies. Transformers \cite{vaswani2017attention} revolutionized natural language processing with context-aware understanding. AlphaGo \cite{silver2017mastering} showcased the ability of reinforcement learning to master strategic games. Despite these spectacular advances, many machine learning algorithms continue to assume that their training and test data distributions are similar. Naturally, such a strong assumption easily falters in real-world scenarios \cite{shen2021towards, zhou2022domain}. In practice, complex and unpredictable discrepancies can arise between training and test data distributions, such as noisy sensory recordings during inference, an abrupt change in weather conditions, evolving user requirements, or entirely new objectives unforeseen at training time. This issue becomes even more pronounced as machine learning algorithms are being applied in more and more practical application-oriented contexts. When the test data distributions start to differ from those encountered during training, an adaptation at test-time would mitigate machine learning failure. In this paper, we provide a comprehensive survey of test-time adaptation studies. 

 
The notion of test-time adaptation was initially posed by Vladimir Vapnik, who stated: ``When solving a problem of interest, do not solve a more general problem as an intermediate step. Try to get the answer that you really need but not a more general one'' \cite{vapnik2006estimation, vapnik2013nature}. A manifestation of this notion are local learning methods \cite{bottou1992local, zhang2006svm} that train algorithms on the neighbors of each test sample before making their prediction. Alternatively, transductive learning \cite{gammerman2013learning, collobert2006large, arnold2007comparative, joachims2012learning} was investigated that incorporates test data information into model training. With the advances of deep learning methods, test-time adaptation \cite{sun2020test, wang2021target} has recently emerged as a new learning method for adaptation at test time. As the name suggests, test-time adaptation achieves the adaptation procedure along with inference to reduce the negative impact of distribution shifts between the training and test data \cite{wang2021tent}. 
During training, the approaches develop general models solely based on the training data distributions, without having access to the target test data. At test time, they focus on adjusting the trained model parameters or test data representations to bridge the gap between the training and test data distributions, thereby enhancing the model's performance and robustness on specific test samples. By facilitating adaptation in conjunction with inference in an online manner, test-time adaptation is computationally efficient and well-suited for scenarios involving online or limited test data, which are so common in real-world use cases.

\begin{figure}[t] 
\centering 
\includegraphics[width=0.45\textwidth]{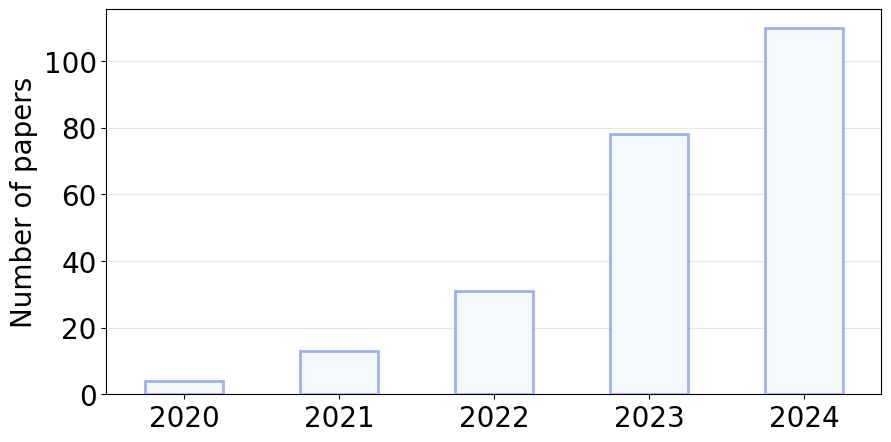}
\vspace{-2mm}
\caption{\textbf{Five-year summary of test-time adaptation research works.} Statistics collected from the seven highest-ranked AI conferences in Google Scholar: CVPR, NeurIPS, ICLR, ICML, ICCV, ECCV, and AAAI. This paper provides a survey on the steadily increasing number of research works. 
} 
\label{fig:numpapera}
\vspace{-4mm}
\end{figure}

The machine learning landscape is witnessing a growing interest in test-time adaptation algorithms, attributed to their ability to handle unseen distribution shifts during test phases and the relatively flexible demands on target data. Figure \ref{fig:numpapera} illustrates this trend, with research efforts on test-time adaptation emerging around 2020, with the pioneering works of Sun \etal \cite{sun2020test} and Wang \etal \cite{wang2021tent}, and expanding in depth and breadth each year. 
%
Two recent surveys also cover test-time adaptation. Liang \etal \cite{liang2023comprehensive} provided a comprehensive overview of test-time adaptation together with source-free domain adaptation. Wang \etal \cite{wang2023search} survey online test-time adaptation methods with their evaluations on Vision Transformers. Unlike previous surveys, which primarily focus on model adaptation techniques, our survey extends the scope to explore how different components of the learning process can be adapted at test time, beyond the model. This wider perspective not only highlights the diversity of approaches within the field but also emphasizes the potential for new research directions that could further enhance the robustness and flexibility of machine learning models across various application domains.
Moreover, methods that go beyond model adaptation offer greater potential for adapting large-scale models during test time, opening up new possibilities for efficient and effective adaptation related to recent advances in multimodal foundation models.

The paper is organized as follows. In Section \ref{background} we provide the problem definition of test-time adaptation and discuss related machine learning problems. In Section \ref{method:w}, we structure our survey by categorizing existing methods into five distinct categories based on what component of the method is adjusted for test-time adaptation: the model, the inference, the normalization, the sample, or the prompt, providing detailed analysis of each. In Section \ref{method:h2p} we further categorize the test-time adaptation methods according to their preparations needed for effective and efficient training. Section \ref{method:h2a}, then moves to deployment and surveys the existing methods according to their update strategies and inference data used at test-time. In Section \ref{app:classification}, we summarize the major evaluation datasets and settings as used by contemporary methods. Section \ref{application} presents the current applications of test-time adaptation algorithms. In Section \ref{futurework}, we provide emerging research opportunities. Finally, we conclude the paper in Section \ref{conclusion}.


%% file: 2_background.tex
\section{Background} 
\label{background}

We first provide the necessary background knowledge on test-time adaptation. In Section \ref{define}, we introduce the problem definition of distribution shifts and test-time adaptation, as well as the notation coventions we use throughout this survey. We then discuss the similarities and differences of test-time adaptation with related learning frameworks in Section \ref{related}.

\begin{figure*}[t] 
\centering 
\centerline{\includegraphics[width=0.99\textwidth]{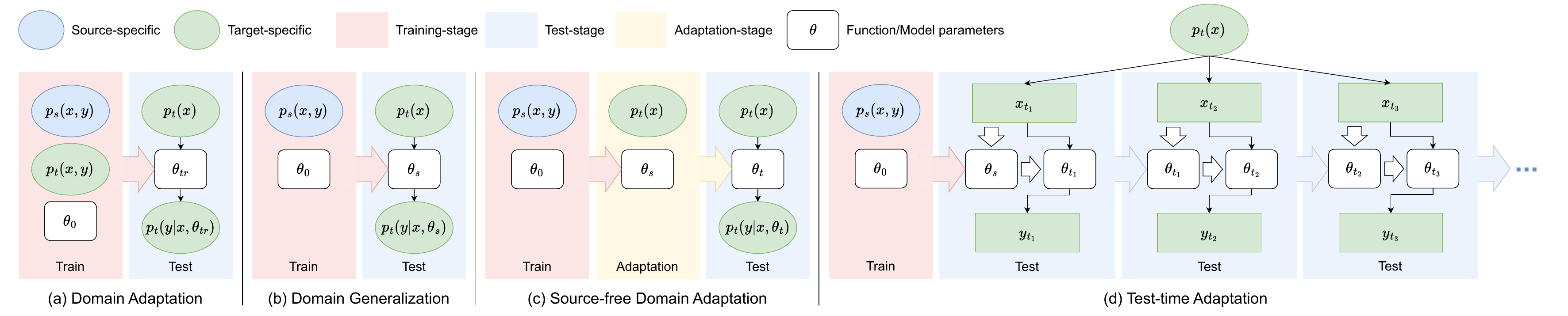}}
\vspace{-2mm}
\caption{
\textbf{Learning frameworks that attack distribution shifts.} (a) \textit{Domain adaptation} addresses domain shifts by accessing both source and target data during training. (b) \textit{Domain generalization} avoids the need for target samples during training, but lacks information on the target distribution during inference. (c) \textit{Source-free adaptation} considers both by introducing an intermediate adaptation stage after source training and before inference. (d) \textit{Test-time adaptation} achieves adaptation on target data along with inference. Test-time adaptation is the focus of this survey, it aims to adjust a source-trained model to target data without having any prior knowledge of the target data before the testing phase.
} 
\label{adapt_settings}
\vspace{-4mm}
\end{figure*} 

\subsection{Problem definition} 
\label{define}

\vspace{.5mm}
\noindent\textbf{Notations.} 
We start from one or more source distributions $N$, defined during training as $\mathcal{S}=\{ p(x_{s_i}, y_{s_i}) \}_{i=1}^N$, and the test-time target distribution $\mathcal{T}=p(x_t, y_t)$, both in the joint space $\mathcal{X} \times \mathcal{Y}$. 
Here $\mathcal{X}$ denotes the input (feature) space and $\mathcal{Y}$ denotes the label space, 
$(x_s, y_s)$ and $(x_t, y_t)$ denote the data-label pairs sampled from the source and target distributions, respectively.
The joint distribution $p(x, y)$ of source and target data can be different.
In the context of test-time adaptation, only the source distribution $\mathcal{S}$ is accessible during training. At test time, only \textit{unlabeled} data $x_t$ from the target distribution is available for adaptation and inference. 
The learned function or model is defined as $f\!\!: \mathcal{X} \rightarrow \mathcal{Y}$ with parameters $\btheta$.
We represent the source-specific model as $f_{\btheta_s}$ and the target model as $f_{\btheta_t}$.

\vspace{.5mm}
\noindent\textbf{Distribution shifts.}
Test-time adaptation focuses on addressing distribution shifts in machine learning algorithms. The fundamental problem is the incongruence between the target and source distributions, i.e., 
\begin{equation}
    p(x_t, y_t) \neq p(x_s, y_s).
\end{equation}
The discrepancy indicates that a source-trained model $f_{\btheta_s}$ can falter when applied to data from the target distribution, leading to predictions that are not as precise and reliable as expected.

By decomposing the joint distribution $p(x, y)$ into $p(x, y) = p(x) p(y|x) = p(y) p(x|y)$, the distribution shifts can be further categorized into four common types, \textit{covariate shifts}, \textit{label shifts}, \textit{conditional shifts}, and \textit{concept shifts} \cite{liu2022deep, xiao2024any}.
\textit{Covariate shifts} ($p(x_t) \neq p(x_s)$, $p(y_t|x_t) = p(y_s|x_s)$) \cite{sun2020test, schneider2020improving} assume the distribution shifts occur only on the input space $p(x)$, while the labels given the input features remain the same. 
In contrast, \textit{label shifts} ($p(y_t) \neq p(y_s)$, $p(x_t|y_t) = p(x_s|y_s)$) focus on the shifts in label space $p(y)$, by assuming the same label-conditioning data distribution.
\textit{Concept shifts} ($p(x_t) = p(x_s), p(y_t|x_t) \neq p(y_s|x_s)$) denote differences in the conditional distributions with the same input distribution, such as noisy labels or different annotation methods \cite{liu2022deep}.
\textit{Conditional shifts} ($p(y_t) = p(y_s), p(x_t|y_t) \neq p(x_s|y_s)$) assume that the label space remains the same, but the distribution of the input samples varies according to the labels, e.g., subpopulation problems \cite{liu2021adversarial, santurkar2020breeds}.
The large majority of test-time adaptation methods focus on covariate shifts \cite{niu2022efficient, schneider2020improving, sun2020test, wang2021tent}, with the distribution shifts coming from the discrepancies in the input data. 
Recently, approaches started to investigate label shifts \cite{park2023label, sun2024beyond, zhou2023ods} and some other distribution shifts \cite{lee2022surgical, xiao2024any}. 
Additionally, some methods have investigated joint-shifts that combine both covariate and label shifts \cite{xiao2024any, park2023label}.

\vspace{.5mm}
\noindent\textbf{Test-time adaptation.}
Given labeled source distributions $\mathcal{S}$ and unlabelled target distributions $\mathcal{T}$, test-time adaptation aims to train a model $f_{\btheta_s}$ only on the source distributions $\mathcal{S}$ and achieve adaptation with the source-trained model $f_{\btheta_s}$ and the target data $\x_t$ to make predictions on $\x_t$ after adaptation. 
The model parameters $\btheta_s$, target data $\x_t$, and even prompts in transformer-based models can be adapted.
Current test-time adaptation approaches adapt each or a combination of these components.
Further note that test-time adaptation is achieved along with inference, in an online or batch-wise manner, without having access to large amounts of target data at each test step.

\subsection{Related problems}
\label{related}

Test-time adaptation draws inspiration from several related machine learning problems that attack distribution shifts, most notably domain adaptation, domain generalization, and source-free domain adaptation, as illustrated in Figure \ref{adapt_settings} and detailed next.

\vspace{.5mm}
\noindent\textbf{Domain adaptation.} 
To handle the shift between source and target distributions, considerable efforts have been devoted to developing methods for domain adaptation \cite{long2015learning,lu2020stochastic,hoffman2018cycada,kumar2010co,tzeng2017adversarial,luo2019taking}.
These methods narrow the gaps between domains using both source and (unlabeled) target samples during training \cite{hoffman2018cycada,kumar2010co,tzeng2017adversarial,luo2019taking}. 
However, the assumption that target data is accessible during training is often invalid in real-world applications. 
Hence, even more challenging settings were proposed in domain adaptation, e.g., few-shot  \cite{motiian2017few} and one-shot domain adaptation \cite{dong2018domain,luo2020adversarial}, but there is still a few or at least one target sample available during training.
By sharp contrast, test-time adaptation assumes not a single target sample is accessible during training.  At test time, the target data and the source-trained model are utilized for adaptation along with evaluation.
An extensive overview of domain adaptation methods can be found in previous survey papers like the ones by Farahani~\etal~\cite{farahani2021brief} and Wang and Deng \cite{wang2018deep}.

\vspace{.5mm}
\noindent\textbf{Domain generalization.} 
Another well-studied problem setting for dealing with distribution shifts is domain generalization, where a model exclusively trained on source data is directly applied to the unseen target samples. In domain generalization, one of the predominant methods is domain invariant learning \cite{ahuja2021invariance, arjovsky2019invariant, muandet2013domain,ghifary2016scatter,motiian2017unified,seo2020learning,zhao2020domain,xiao2021bit,mahajan2021domain,nguyen2021domain, li2018domainb, phung2021learning,shi2021gradient}.
These methods learn an invariant feature space on source distributions and hope that it will generalize well to the unseen target distributions. Another widely used methodology is domain augmentation \cite{shankar2018generalizing,volpi2018generalizing,qiao2020learning, zhou2020learning,zhou2021mix, yao2022improving}, which generates more source domain data to learn more robust feature representations during training. 
Meta-learning-based methods have also been studied for domain generalization \cite{li2018metalearning,balaji2018metareg,dou2019domain,bui2021exploiting,du2021hierarchical}. They learn the ability of generalization by mimicking distribution shifts within the source distributions during training. 
Detailed categorizations and analysis of domain generalization methods are provided in survey papers by Zhou \etal~\cite{zhou2022domain} and Wang \etal~\cite{wang2022generalizing}.
All domain generalization methods utilize only source data during training. Since the source-trained model is directly deployed on the target distributions, no target information is considered for the deployment.
Therefore, performance will degrade when the target distribution is far off from the source distribution.
By contrast, test-time adaptation considers the target information by adapting the source-trained model to target distributions during inference. By doing so achieving a more robust and stable performance across different distribution shifts.

\vspace{.5mm}
\noindent\textbf{Source-free domain adaptation.}
To avoid the requirement of having target data during training, while considering the target information for good generalization across distribution shifts, source-free adaptation \cite{fang2022source, jing2022variational, kundu2020universal, liang2020we, yu2023comprehensive} is proposed.
As shown in Figure \ref{adapt_settings}, source-free domain adaptation methods first train their model on source distributions, then adapt the source-trained model on the entire target set before evaluation.
During adaptation and testing, no source data is available. Source-free domain adaptation is commonly achieved by model fine-tuning \cite{lee2022confidence, liang2020we, jing2022variational, yang2021exploiting, yang2022attracting, yi2023source} and data generation \cite{ding2022source, kundu2022balancing, li2020model}.
Recent survey papers provide detailed overviews of source-free adaptation methods \cite{fang2022source, liang2023comprehensive, yu2023comprehensive}.
Compared with source-free domain adaptation, test-time adaptation methods adapt a source-trained model in an online manner. In other words, adaptation is achieved during inference or along with inference. 
Thus, test-time adaptation avoids the additional adaptation procedure for iterative model fine-tuning before inference, which is more computationally efficient. Because of the online nature, test-time adaptation also relies less on the target data leading to better generalization, especially for diverse target distributions.

%% file: 3_methods_what.tex

\section{What to adapt}
\label{method:w}


Given a source-trained model and new target data distributions that appear at test time, the goal of test time adaptation is to fit the source-trained model to the unseen target data. In this section, we categorize the existing methods from the literature into five main categories, based on what they adapt: the model, the inference, the normalization, the sample, or the prompt. In the following sections, we present the details of each category along with an overarching equation and the corresponding methods. The adaptation component in each equation is highlighted in red.


\subsection{Model adaptation}
\label{sec:model-adapt}


\begin{figure*}[t]
\centering
\includegraphics[width=0.9\linewidth]{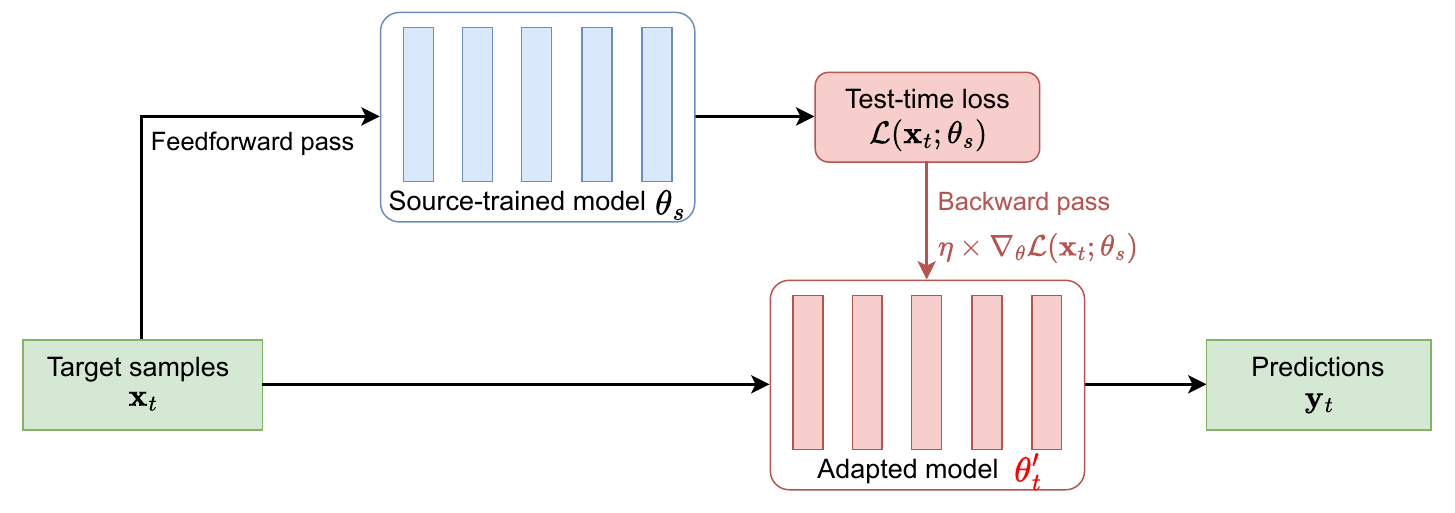}
\vspace{-2mm}
\caption{
\textbf{Model adaptation.} These methods gradually update their source-trained model by backpropagating a self-trained loss on target test data. 
}
\label{fig3:modeladapt}
\vspace{-5mm}
\end{figure*}

An intuitive way of test-time adaptation is by adjusting the source-trained model parameters $\btheta_s$ according to the target data $\x_t$. By doing so, obtaining the target-specific model parameters $\btheta_t$. Most model adaptation methods adjust their source-trained model by fine-tuning the model parameters through unsupervised losses on the target test data, see Figure \ref{fig3:modeladapt}. 

Given an unsupervised loss $\mathcal{L}(\x_t; \btheta)$ on target data, the target-specific model $\btheta_t$ and the prediction $\y_t$ is obtained by
\begin{equation}
    \textcolor{red}{\btheta_t'} = 
    \min_{\btheta} \mathcal{L}(\x_t; \btheta_s) = 
    \btheta_s + \eta \nabla_{\btheta} \mathcal{L}(\x_t; \btheta_s), ~~~ \y_t = f_{\textcolor{red}{\btheta_t'}}(\x_t),
\end{equation}
where $\nabla_{\btheta} \mathcal{L}(\x_t; \btheta_s)$ denotes the parameter gradient of unsupervised loss $\mathcal{L}$ and $\eta$ denotes the learning rate of backpropagation. 
The adaptation of $\btheta_t'$ requires iterative optimization. We categorize the model adaptation methods further according to their unsupervised loss functions.

\vspace{.5mm}
\noindent
\textbf{Auxiliary self-supervision.} 
Sun \etal~\cite{sun2020test} pioneered model adaptation by self-supervision.
%
%
Their test-time training procedure introduces an auxiliary self-supervised task with the loss function $\mathcal{L}^{a} (\x)$. The model parameters $\btheta$ are split into the main-task parameters $\btheta^m$ and the shared parameters $\btheta^e$. 
The auxiliary task also has its own task-specific parameters $\btheta^{a}$. 
The standard empirical risk minimization $\min_{\btheta} \mathcal{L}_s(\x_s, \y_s; \btheta)$ during source training is then changed to:
\begin{equation}
    \min_{\btheta^e, \btheta^m, \btheta^a} \mathcal{L}^m(\x_s, \y_s; \btheta^e, \btheta^m) + \mathcal{L}^a(\x_s; \btheta^e, \btheta^a),
\end{equation}
where $\mathcal{L}^m$ and $\mathcal{L}^a$ denote the loss function of the main task and auxiliary task, respectively.
After training, the source model consists of $\btheta_s = (\btheta_s^e, \btheta_s^m, \btheta_s^a)$.
At test time, the main-task model $\btheta^m$ is fixed while the shared model $\btheta^e$ is fine-tuned by minimizing the auxiliary task loss on the target data $\x_t$, which is formulated as:
\begin{equation}
    \btheta^e_t = \min_{\btheta_e} \mathcal{L}^a(\x_t; \btheta^e_s, \btheta^a_s).
\end{equation}
Finally, the model makes predictions on the target data using the target model $\btheta_t' = (\btheta^e_t, \btheta_s^m)$.
Since the model is optimized only by the auxiliary task during training, it is necessary to avoid model overfitting on the auxiliary task. Therefore, the training stage is altered to jointly optimize the model by the main task and the auxiliary task, expecting to make the adaptation of the auxiliary task compatible with the main task \cite{sun2020test,liu2021ttt++}. 

To further enhance model adaptation by auxiliary self-supervision, several auxiliary tasks have been proposed. Varsavsky \etal~\cite{varsavsky2020test} design an adversarial loss and augmentation consistency regularization.
TTT++ \cite{liu2021ttt++} introduces SimCLR \cite{chen2020simple} as the self-supervised auxiliary task. They also propose an online feature alignment strategy, which aligns the feature distributions at test time with the one at training time to avoid overfitting to the auxiliary task. TTT-MAE \cite{gandelsmantest2022test} adapts the model for each test input using masked autoencoders \cite{he2022masked} for self-supervision. Mate \cite{mirza2023mate} utilizes a 3D self-supervised reconstruction loss for model adaptation in a 3D classification setting.
Diffusion-TTA \cite{prabhudesai2023diffusion} utilizes the feedback of generative models to adapt the discriminative model.
NC-TTT \cite{osowiechi2024nc} utilizes the discrimination of noisy features for adaptation.
MT3 \cite{bartler2022mt3} relies on BYOL \cite{grill2020bootstrap} as the auxiliary task for test-time training.
Both Sain \etal~\cite{sain2022sketch3t} and Liu \etal~\cite{liu2023meta} supervise their model adaptation procedure by a self-supervised image reconstruction objective. 
ClusT3 \cite{hakim2023clust3} introduces an unsupervised loss, maximizing the mutual information between the features of different layers.

Besides innovations on the self-supervision, many methods enforce better compatibility and cooperation between the main and auxiliary tasks.
Li \etal~\cite{li2021test} introduce a transformer to better couple the relationship between the main and auxiliary tasks for human pose estimation.
ActMAD \cite{mirza2022actmad} proposes activation matching for more fine-grained supervision.
OWTTT \cite{li2023robustness} extends model adaptation by self-supervision on open-world data, supervised by clustering and distribution alignment at test time.
Several methods \cite{alet2021tailoring, chen2022ost, choi2021test, min2023meta} learn to enhance the cooperation between main and auxiliary tasks through meta-learning \cite{finn2017model, finn2018probabilistic, vanschoren2018meta, wu2024test}. Tailoring \cite{alet2021tailoring} trains its model to perform well on the task loss after adaptation using the unsupervised loss. MT3 \cite{bartler2022mt3} uses the same approach to achieve model adaptation on a single unlabeled image.
Min \etal~\cite{min2023meta} deploy the meta-learning strategy into optical flow networks.

Model adaptation by auxiliary self-supervision requires careful design of the auxiliary tasks and their coupling with the main task to avoid overfitting. Often necessitating both main and auxiliary losses, leading to an alternating training stage with extra computational costs.
As an alternative, entropy minimization of model predictions is proposed to achieve test-time model adaptation without any auxiliary task.



\vspace{.5mm}
\noindent
\textbf{Entropy minimization.}
Model adaptation by entropy minimization does not introduce any change in the source-training procedure, but only adapts the source-trained model to the target data at test time.
Wang \etal ~\cite{wang2021tent} propose Tent, which introduces the concept of fully test-time adaptation. It fine-tunes the source-trained model parameters on target distributions directly by minimizing the entropy of the model predictions.
\begin{equation}
    \btheta_t' = \min_{\btheta} \mathcal{L}(\x_t, \btheta_s) = \min_{\btheta} H(\hat{\y}_t),
\end{equation}
where the entropy $H(\hat{\y}_t)$ is conducted on the source-model predictions on the target sample $\hat{\y}_t = f_{\btheta_s} (\x_t)$.
Different from auxiliary self-supervised methods that split the main task parameters $\btheta$ into shared parameters $\btheta_e$ and specific parameters $\btheta_m$ and only update $\btheta_e$ at test time, Tent can adjust all parameters $\btheta$ by test data. Hence, the loss function and update parameters directly relate to the main task. 

Inspired by Tent, many methods enhance entropy minimization for better adaptation in specific use cases \cite{marsden2023universal, niu2022efficient, zhang2021memo, chen2024towards}.
MEMO \cite{zhang2021memo} augments each target sample and achieves adaptation by minimizing the marginal entropy of the predictions for different augmentations.
EATA \cite{niu2022efficient} finds that adaptation on high entropy test samples can hurt performance. They propose adaptive entropy minimization to adapt low-entropy samples only.
DeYO \cite{lee2024entropy} introduces a probability difference to measure the difference between
predictions before and after augmentation as an auxiliary metric for sample selection and weighting.
SoTTA \cite{gong2023sotta} removes the low-confidence inputs and large gradients for robust adaptation on noisy test data.
Lee \etal~\cite{lee2024stationary} proposes Bayesian filtering to combine test and training during online model adaptation.
CMF \cite{lee2024continual}  utilizes a Kalman filter
to strike a balance between model adaptation and information retention.
Choi \etal~\cite{choi2022improving} complement the entropy minimization with a mean entropy maximization, as well as an auxiliary classification task based on the nearest source prototype classifier. 
Lee \etal~\cite{lee2023towards} select target samples before entropy minimization, where the samples are filtered out if the adapted confidence is lower than the original one.
DomainAdaptor \cite{zhang2023domainadaptor} introduces generalized entropy minimization, which includes a temperature scaling for the original entropy loss.  
AETTA \cite{lee2024aetta} proposes prediction disagreement with dropout inferences for more robust accuracy estimation of test-time adaptation methods. 
STAMP \cite{yu2024stamp} achieves adaptation over a stable memory bank with self-weighted entropy minimization.
Bar \etal~\cite{bar2024protected} matches the distribution of test entropy values to the source ones to adaptively update model parameters.

Entropy minimization enables an unsupervised model adaptation that is highly related to the main task, without alternating the training stage and the need to design auxiliary self-supervised tasks.
However, due to distribution shifts, the source-model prediction on target data can be incorrect. In this case, entropy minimization methods may lead to error accumulation during model adaptation.

\vspace{.5mm}
\noindent
\textbf{Pseudo-labeling.}
Besides entropy minimization, pseudo-labeling \cite{galstyan2008empirical,lee2013pseudo} is also widely used for model adaptation \cite{rusak2021if, zancato2023train}. 
Intuitively, pseudo-labeling is similar to entropy minimization, as both tend to optimize the source-trained model by maximizing the confidence of the model predictions on the target data. The key difference is that pseudo-labeling provides more explicit supervision and is easier to refine. Pseudo-labeling model adaptation is formulated as:
\begin{equation}
\label{eq: pseudolabel}
    \tilde{\y}_t = \arg\max_{\hat{\y}_t} (\hat{\y}_t = f_{\btheta_s}(\x_t)), ~ \btheta_t' = \min_{\btheta} \mathcal{L}^{ce}(\x_t, \tilde{\y}_t; \btheta),
\end{equation}
where $\tilde{\y}_t$ denotes the pseudo-labels obtained from the source model predictions $\hat{\y}_t$.
$\mathcal{L}^{ce}$ denotes the cross-entropy loss between model predictions and pseudo-labels.
The pseudo-labels $\tilde{\y}_t$ can be either hard (i.e., one-hot) or soft (i.e., continuous).
Due to distribution shifts between the source and target data, the pseudo-labels of the unseen target data can be inaccurate.
Therefore, it is necessary to enhance and refine the pseudo-labels \cite{wang2022continual, liang2020we, wang2023soda, yu2024domain, ma2024improved}.


\begin{figure*}[t]
\centering
\includegraphics[width=0.9\linewidth]{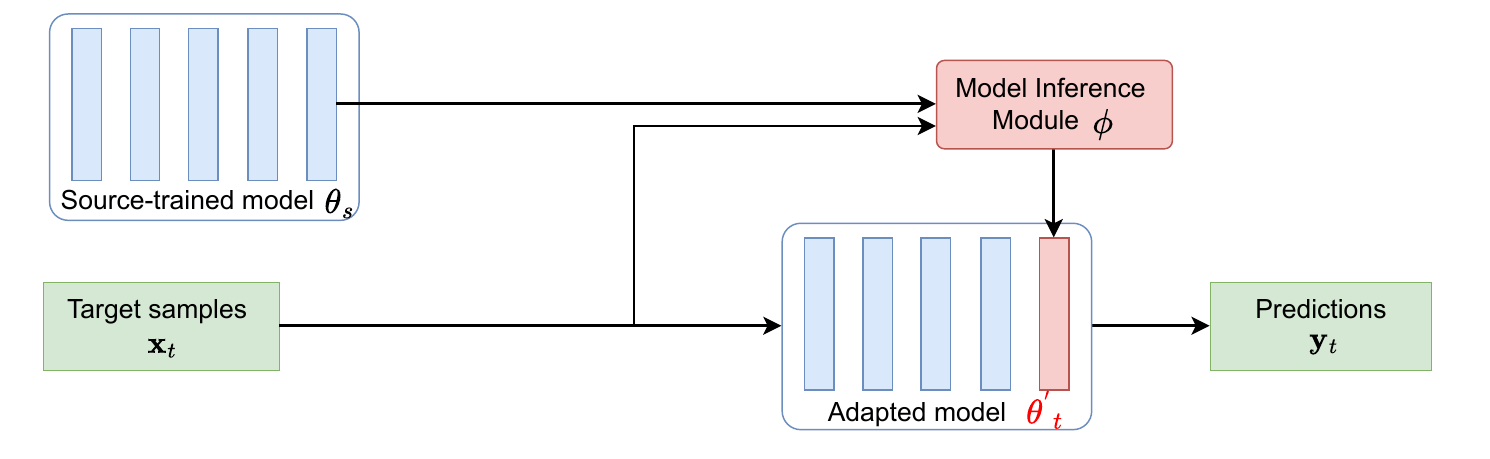}
\vspace{-2mm}
\caption{
\textbf{Inference adaptation.} These methods generate model parameters by an auxiliary model-inference module in a single-forward pass, without any back-propagation. 
}
\label{fig3:modelinfer}
\vspace{-5mm}
\end{figure*}

Based on teacher and student networks, Rusak \etal~\cite{rusak2021if} explore hard pseudo-labeling, soft pseudo-labeling, and entropy minimization. The work further proposes robust pseudo-labeling, 
which replaces the cross-entropy loss $\mathcal{L}^{ce}$ in eq. (\ref{eq: pseudolabel}) with the generalized cross-entropy loss 
to solve the problem of training stability and hyperparameter sensitivity in the common pseudo-labeling methods.
Wang \etal~\cite{wang2021target} initialize their student model by contrastive learning from scratch on the target data and fine-tune it with the pseudo-labels generated by the teacher model. 
Differently, TeST \cite{sinha2023test} initializes their student model by the source-trained model and fine-tunes it with the pseudo-labels and entropy minimization.
TeSLA \cite{tomar2023tesla} introduces flipped cross-entropy that uses the soft pseudo-labels from the teacher network together with an entropy maximization of the predictions. 

Naturally, the pseudo-labels can be refined with the information of the target samples \cite{wang2022continual,chen2022contrastive,jang2022test}. 
CTTA \cite{wang2022continual} generates pseudo-labels by a weighted-averaged teacher model together with the augmentation-averaged pseudo-labels.
Chen \etal~\cite{chen2022contrastive} propose Contrastive TTA with an online pseudo-label refinement that aggregates knowledge of the nearby target samples and fine-tunes the model with both pseudo-label cross-entropy and self-supervised contrastive learning.
TAST \cite{jang2022test} also generates pseudo-labels using the nearest neighbors from a set composed of previous test data.
In addition to refining pseudo-labels with neighboring target samples, 
Litrico \etal~\cite{litrico2023guiding} further reweight the loss based on the reliability of the pseudo-labels.
Ambekar \etal~\cite{ambekarvariational} consider the uncertainty of the pseudo-labels by formulating adaptation as a probabilistic inference problem where pseudo-labels act as latent variables.
Goyal \etal~\cite{goyal2022test} find that different loss functions perform best for the classifier with different training losses.
They propose conjugate pseudo-labels to specify a good adaptation loss at test time for any training loss function. PROGRAME \cite{sun2024program} constructs a graph using prototypes and test samples to generate more reliable pseudo-labels.
WATT \cite{osowiechi2024watt} averages the updated model parameters in different optimization steps to facilitate test-time adaptation.

Similar to entropy minimization, pseudo-labeling methods also suffer from error accumulation due to inaccurate pseudo-labels. Therefore, obtaining more accurate labels and dealing with noisy pseudo-labels is essential. Moreover, supervisions like entropy minimization and pseudo-labeling are mainly conducted on the output space. Therefore, the methods are often proposed for specific tasks like classification. Naturally, there are also methods that operate on the intermediate feature space.

\vspace{.5mm}
\noindent
\textbf{Feature alignment.}
Common unsupervised objective functions on the feature space are feature alignment and consistency \cite{lin2022video, fleuret2021test, jung2022cafa,  su2022revisiting, su2023revisiting, wang2023feature, li2024bi, modi2024asynchronous}.
The model optimized by the feature alignment or consistency objective functions is formulated as:
\begin{equation}
\label{eq: align}
    \btheta_t' = \min_{\btheta} \mathcal{L}^{align}(\z_t, \z_a; \btheta),
\end{equation}
where $\z_t$ and $\z_a$ denote the feature representations of the test sample $\x_t$ and auxiliary samples $\x_a$. The auxiliary samples $\x_a$ can be obtained from the neighboring target data \cite{wang2023feature}, source data \cite{su2022revisiting}, and augmented data \cite{nguyen2023tipi, fleuret2021test, lin2022video}.
$\mathcal{L}^{align}$ denotes the alignment or consistency objective functions, such as the MSE loss or KL divergence.
CAFA \cite{jung2022cafa} designs a class-aware feature alignment function to learn test data in a class-discriminative manner.
Inspired by unsupervised domain adaptation methods, Su \etal~\cite{su2022revisiting} propose anchored clustering to align the source and target clusters and improve the clustering by pseudo-label filtering and iterative updating at test time.
Fleuret \etal~\cite{fleuret2021test} introduce an augmentation consistency loss to enforce consistent predictions of the augmented target samples.
Kang \etal~\cite{kang2023leveraging} regularize the adaptation by matching the proposed proxies between training and test data.
TIPI \cite{nguyen2023tipi} introduces a transformation invariance regularizer as the objective function for test-time adaptation. 
Wang \etal~\cite{wang2024distribution} proposes distribution alignment to guide the test feature distribution back to the source for continual test-time adaptation. 
Compared with entropy minimization and pseudo-labeling that are specific for classification tasks, feature alignment can be used for a wider range of tasks beyond classification. However, it can also lead to suboptimal adaptation since the decoder of the model is difficult to adapt.

\vspace{.5mm}
\noindent
\textbf{Discussion.}
Model adaptation methods achieve considerable improvements in out-of-distribution generalization with enough unlabeled test data and computational resources.
However, the methods also have some challenges and weaknesses.
Since the methods need incremental model updates, model adaptation at test time is computationally demanding \cite{xiao2022learning, gao2022back}.
Moreover, the methods are known to be unstable in various test-time scenarios \cite{niu2022efficient, wang2022continual, niu2023towards}. 
For example, model adaptation may fail or even harm the model robustness when experiencing continually changing distributions at test time \cite{wang2022continual, brahma2022probabilistic}, or when experiencing a mixture of multiple distributions \cite{xiao2022learning}, or for very small batch sizes \cite{zhang2021memo, niu2023towards}. Wu \etal~\cite{wu2023uncovering} further find that model adaptation methods are vulnerable to malicious data.
To address these problems, recent methods focus on new settings like continual test-time adaptation \cite{brahma2022probabilistic, niu2022efficient, wang2022continual} and limited-data adaptation \cite{niu2023towards, zhang2021memo}, which we will further discuss in Section \ref{method:h2a}.

\begin{figure*}[t]
\centering
\centering
\includegraphics[width=0.9\linewidth]{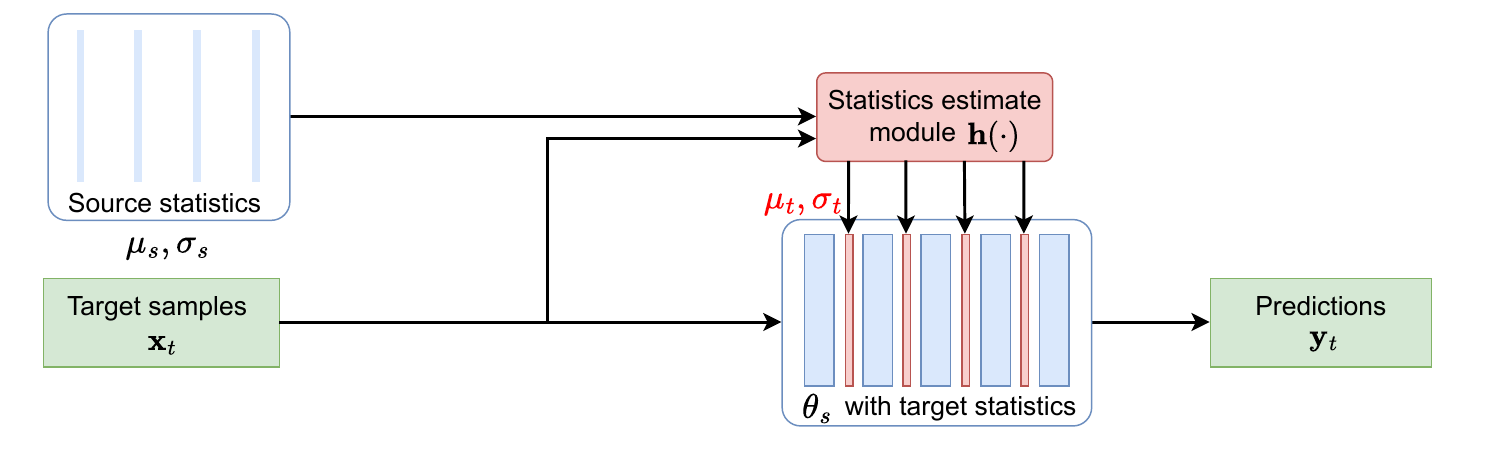} 
\vspace{-2mm}
\caption{
\textbf{Normalization adaptation.} These methods adjust their normalization statistics while fixing source parameters.
}
\label{fig4:normada}
\vspace{-5mm}
\end{figure*}

\subsection{Inference adaptation}
%
%
To avoid the model adaptation requirements of having a large number of target samples and iterative compute, inference adaptation methods estimate model parameters with a small number of samples using just a feedforward pass at test time \cite{iwasawa2021test, xiao2022learning, dubey2021adaptive}, as illustrated in Figure \ref{fig3:modelinfer}.
For these methods the target-specific model parameters are obtained by:
\begin{equation}
\textcolor{red}{\btheta_t'} = \bphi(\x_t, \btheta_s), ~~~ \y_t = f_{\textcolor{red}{\btheta_t'}}(\x_t),
\end{equation}
where $\bphi$ denotes the model-inference module. The model inference procedure usually requires the information from the target data $\x_t$ and source-trained model parameters $\btheta_s$. 
Since directly inferring the model parameters is difficult, it may require additional information from the target data.
According to the amount of required test data, we divide them into batch-wise and sample-wise inference methods.

\vspace{.5mm}
\noindent
\textbf{Batch-wise inference.} 
Batch-wise inference adaptation requires batches of target samples to update the model at test time.
To infer their target-specific model, Dubey \etal~\cite{dubey2021adaptive} introduce a domain-specific function that generates domain embeddings through an MLP network.
Differently, T3A \cite{iwasawa2021test} proposes a parameter-free inference module. The method utilizes averaged features to online adjust the classifiers according to pseudo-labels at test time. 
To avoid error accumulation, they filter the unreliable pseudo-labeled data with high entropy.
AdaNPC \cite{zhang2023adanpc} also achieves a non-parametric inference module that infers the classifier by storing the target information in an online updated memory at test time.
The online update method is also utilized in continuous-time Bayesian neural network \cite{huangextrapolative}, which infers model parameters by a particle filter differential equation.
Zhou \etal~\cite{zhou2023test} introduce distribution normalization on multimodal features to improve similarity measurement between modalities at test time.
Batch-wise inference adaptation avoids the computational cost of backpropagation for obtaining target-specific models. However, it requires batches of target samples to be sampled from the same distribution, which may be difficult to obtain and assure in real-world applications.

\vspace{.5mm}
\noindent
\textbf{Sample-wise inference.}
To achieve inference adaptation with fewer target samples, Xiao \etal~\cite{xiao2022learning} propose to generate a sample-specific classifier for each target sample, where the inference module $\bphi$ is an MLP network trained by variational inference. 
DIGA \cite{wang2023dynamically} infers a prototypical classifier for each instance in a semantic segmentation task.
Sun \etal~\cite{sun2022dynamic} meta-learn the module $\bphi$ for inferring instance-specific model parameters.
VoP \cite{kim2022variational} introduces variational inference in the inference module to achieve personalized test-time adaptation.
They estimate the model parameters on the fly based on the personality of a small amount of personal data.
RNA \cite{yeo2023rapid} trains a small inference network to predict parameters for modulating the original features.
FedIns \cite{feng2023towards} obtains the instance-adaptive model by adaptively selecting the best-matched subsets of learned feature pools in federated learning.
TDA \cite{karmanov2024efficient} introduce dynamic caches that maintain few-shot test features and pseudo labels to enhance the model prediction at test time. 
BoostAdapter \cite{zhang2024boostadapter} leverages a light-weight key-value memory.
DPE \cite{zhang2024dual} infers prototypes from both textual and visual modalities from CLIP and optimizes prototypes using learnable residual parameters for each test sample.
Sample-wise inference adaptation reduces the requirement for large amounts of data, making it much more data-efficient at test time. 
However, the sample-wise inference is more difficult due to the limited amount of information in one sample, which usually requires specifically designed training strategies like meta-learning \cite{sun2022dynamic, xiao2022learning}.

\vspace{.5mm}
\noindent
\textbf{Discussion.}
Compared to model adaptation, inference adaptation methods are more efficient since they achieve model generation and model updates by a single feedforward pass of the target data, without backpropagation. However, most inference adaptation methods require additional designs of the source-training stage, which may be not as convenient as model adaptation methods.
Moreover, due to the large number of network model parameters, the model inference methods often can only focus on a subset of the parameters, e.g., the classifier layer \cite{iwasawa2021test, xiao2022learning, dubey2021adaptive}. As different distribution shifts are influenced by different network layers \cite{lee2022surgical}, it is difficult for these methods to achieve good adaptation.

\subsection{Normalization adaptation}

Since adjusting model parameters can be computationally expensive and training dependent, normalization adaptation methods focus on adjusting only the normalization statistics of the widely used batch normalization layers \cite{ioffe2015batch}.
batch normalization is known to reduce the internal covariate shift by normalizing the input features of subsequent layers by $\hat{\x}_s = \frac{\x_s - \mu_s}{\sqrt{\sigma_s^2 + \epsilon}}$, where $\mu_s$ and $\sigma_s^2$ are the expectation and variance of the mini-batch of samples during source training. 
Since the target data $\x_t$ and source statistics $\mu_s, \sigma_s$ are mismatched, the normalization adaptation methods estimate the target statistics $\mu_t, \sigma_t$ to normalize the target features while fixing the source-trained model parameters, as shown in Figure \ref{fig4:normada}.
The adaptation process is formulated as:
\begin{equation}
    \textcolor{red}{\mu_t}, \textcolor{red}{\sigma_t} = \mathbf{h}(\x_t, \mu_s, \sigma_s), ~~~ 
    \textcolor{red}{\hat{\x}_t} = \frac{\x_t - \textcolor{red}{\mu_t}}{\sqrt{\textcolor{red}{\sigma^2_t} + \epsilon}}, ~~~
    \y_t = f_{\btheta_s}(\textcolor{red}{\hat{\x}_t}), 
\end{equation}
where $\mathbf{h}(\cdot)$ denotes the target statistics estimation module. $\mathbf{h}(\cdot)$ can be either parametric or non-parametric, and the inputs of the module are usually the target features $\x_t$, sometimes together with the source statistics $\mu_s, \sigma_s$.

Some model adaptation methods from Section~\ref{sec:model-adapt} also change the batch normalization statistics to improve performance, e.g., Tent and some subsequent methods \cite{wang2021tent, yang2022test, su2024towards} estimate the target mini-batch statistics during fine-tuning the affine parameters of the batch normalization layers.
In this section, we focus on methods that only adjust the normalization statistics, without altering their model parameters. 
We divide current methods into three types according to the target statistics estimation method: calculating target statistics directly, combining source and target statistics, and target statistics inference.

\vspace{.5mm}
\noindent
\textbf{Target statistics.}
To make the target data fit source-trained model parameters, the target features $\hat{\x}_t$ at each layer of the deep neural network need to be normalized at test time, similarly to the source features during training.
One of the most straightforward methods is to directly estimate the statistics of the target data in the batch normalization layers at test time \cite{nado2020evaluating, li2016revisiting}.
Nado \etal~\cite{nado2020evaluating} propose prediction-time batch normalization to recompute the Batch Norm statistics, i.e., $\mu_t$ and $\sigma_t^2$ for each test batch. 
The method enhances calibration under covariate shifts.
Kaku \etal~\cite{kaku2020like} propose adaptive normalization that also utilizes the target batch normalization statistics to estimate the feature statistics of each test instance.
In addition to recomputing batch normalization statistics, ARM \cite{zhang2021adaptive} further changes the training stage by sampling each training batch from the same domain. 
This mimics the statistics computation at test time and boosts the performance by meta-learning. 
MedBN \cite{park2024medbn} replaces the means with medians when computing BN statistics to defend data poisoning attacks.

Replacing the source statistics directly with the target ones can alleviate the covariate shifts between source and target distributions with sufficient target samples.
However, batch normalization always requires large batch sizes or moving averages of large numbers of samples for accurate estimation of the statistics. 
When the target samples are insufficient or from various distributions, the estimated target statistics will not be able to represent the target distribution well, leading to unstable prediction and performance degradation.
UnMix-TNS \cite{tomar2024mixing} re-calibrates the statistics for each test sample by mixing it with multiple online statistics, which are online updated by the most similar samples.
Kaku \etal~\cite{kaku2020like} replace the batch normalization with instance normalization \cite{ulyanov2016instance}, which considers the instance-wise statistics and is widely adopted in estimating style \cite{huang2017arbitrary}, task-specific \cite{bronskill2020tasknorm}, or domain-specific \cite{seo2020learning} information. 
Gong \etal~\cite{gong2022note} proposes instance-aware batch normalization, which corrects the training batch normalization statistics by the instance normalization statistics when the instance-wise statistics of the target sample are significantly different from the training one. Instance normalization addresses the requirement for large numbers of i.i.d target samples of batch normalization, but it makes the features less discriminative over classes \cite{seo2020learning}. 
Therefore, it is unsuitable for complex tasks with a large number of categories.
In addition, large distribution shifts between the source and target domains can result in a mismatch between target batch statistics and source parameters, which will potentially deteriorate the discriminative structures and perturb the predictions \cite{you2021test}.
Therefore, methods that combine source and target statistics are proposed to estimate more stable normalization statistics for domain shifts.

\begin{table}[t]
\caption{\textbf{Statistic estimation methods for normalization adaptation.} The methods generate test-time specific statistics for normalized target features to suit the source-trained parameters.
$X_t = \{ \x_t^m \}^M_{m=1}$ denotes a batch of target features with batch size $M$. $\x_t$ denotes the target feature with the size of $H \times W \times C$. 
Methods using target statistics directly leverage batch- or instance-level statistics of target data. Statistics combination methods aggregate both source and target data statistics to balance source-learned knowledge and target-specific information. Statistics inference methods train a network to infer target-specific statistics.}
\label{tab: norm}
\vspace{-2mm}
\resizebox{0.99\columnwidth}{!}{%
\setlength\tabcolsep{4pt} 
\begin{tabular}{ll}
\toprule
        Adapted statistics used at test time & Equation examples\\ 
        \midrule
        \rowcolor{mColor2}
        \multicolumn{2}{l}{\textbf{Target statistics} } \\
        \multirow{2}*{Target batch statistics, e.g.,\cite{zhang2021adaptive, nado2020evaluating} } & $\mu_t = \frac{1}{MHW} \sum_{m,h,w} X_t$, \\
        ~ & $\sigma_t^2 = \frac{1}{MHW} \sum_{m,h,w} (X_t - \mu_t)^2$ \\ 
        \midrule
        \multirow{2}*{Target instance statistics, e.g., \cite{kaku2020like, gong2022note}} & $\mu_t = \frac{1}{HW} \sum_{h,w} \x_t$,  \\
        ~ & $\sigma_t^2 = \frac{1}{HW} \sum_{h,w} (\x_t - \mu_t)^2$ \\ \midrule
        \rowcolor{mColor2}
        \multicolumn{2}{l}{\textbf{Statistics combination}} \\
        \multirow{2}*{Batch-wise combination, e.g., \cite{schneider2020improving, you2021test}} & $\bar{\mu} = \alpha \mu_t + (1-\alpha) \mu_s$,\\
        ~ & $\bar{\sigma} = \alpha \sigma_t + (1-\alpha) \sigma_s$ \\ \midrule
        \multirow{2}*{Moving-average combination, e.g., \cite{hu2021mixnorm, mirza2022norm}} & $\bar{\mu}_{k+1} = (1-\lambda) \bar{\mu}_k + \lambda \mu_t$,\\
        ~ & $ \bar{\sigma}_{k+1} = (1-\lambda) \bar{\sigma}_k + \lambda \sigma_t$ \\ \midrule
        \rowcolor{mColor2}
        \multicolumn{2}{l}{\textbf{Statistics inference}} \\
        Inferred statistics, e.g., \cite{du2020metanorm, jiang2023domain} & $(\mu_t, \sigma_t) = g_{\bpsi}(\x_t)$ \\ 
\bottomrule
\end{tabular}
}
\vspace{-4mm}
\end{table}

\begin{figure*}[t]
\centering
\centering
\includegraphics[width=0.9\linewidth]{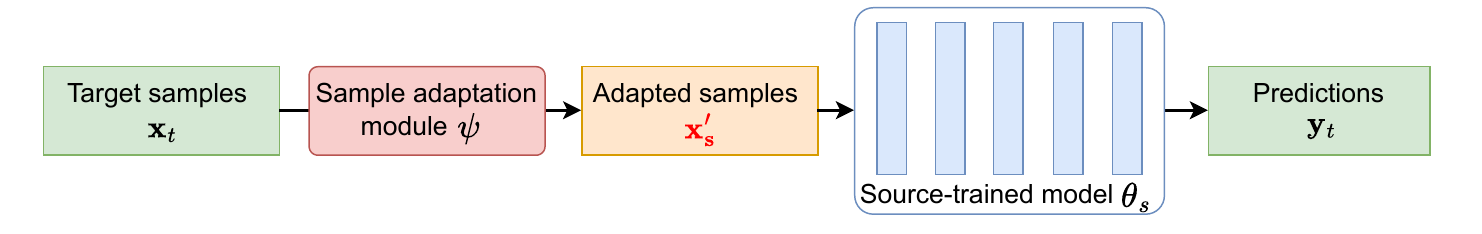} 
\vspace{-2mm}
\caption{
\textbf{Sample adaptation.} These methods adjust the target samples to source distributions, usually by generative models.
}
\label{fig4:sampleada}
\vspace{-5mm}
\end{figure*}

\vspace{.5mm}
\noindent \textbf{Statistics combination.}
One of the most common methods for combination is the weighted sum of the source and target statistics \cite{yao2023generalized, schneider2020improving, lim2023ttn}.
Schneider \etal~\cite{schneider2020improving} introduce a hyperparameter $M$ to combine and trade off the source and target statistics when the number of available target samples $n$ is too small. 
The combined statistics is obtained by $\bar{\mu} = \frac{M}{M + m} \mu_s + \frac{m}{M+ m} \mu_t$, $\bar{\sigma}^2 = \frac{M}{M + m} \sigma^2_s + \frac{m}{M + m} \sigma^2_t$.
Similarly, You \etal~\cite{you2021test} propose $\alpha$-BN to calibrate the batch normalization statistics as $\bar{\mu} = \alpha \mu_t + (1-\alpha) \mu_s$, $\bar{\sigma} = \alpha \sigma_t + (1-\alpha) \sigma_s$, where $\alpha$ is also a hyperparameter.
To extend the method to the case with only a single target sample, SITA \cite{khurana2021sita} obtains $\mu_t$ and $\sigma_t$ of the aforementioned equations from various augmentations of the target sample. 
They achieve a hyperparameter-free combination algorithm by selecting prior hyperparameter values using the prediction entropy.
TEMA \cite{su2024unraveling} proposes an exponential moving average method to combine training and test statistics.
TTN \cite{lim2023ttn} converts the hyperparameter to the learnable parameter. The parameter $\alpha$ is initialized by a gradient distance score and optimized during the proposed post-training phase. 
At test time, the batch normalization statistics are obtained by $\bar{\mu} = \alpha \mu_t + (1-\alpha) \mu_s$, $\bar{\sigma}^2 = \alpha \sigma^2_t + (1-\alpha) \sigma^2_s + \alpha (1-\alpha) (\mu_t - \mu_s)^2$ with the learned $\alpha \in [0, 1]$.

To achieve a stable and robust test-time adaptation, some methods estimate their batch normalization statistics in an online manner at test time \cite{yang2022test, hu2021mixnorm, mirza2022norm}.
Like during training, these methods estimate the statistics through a moving average, e.g., $\bar{\mu}_{k+1} = (1-\lambda) \bar{\mu}_k + \lambda \mu_t, \bar{\sigma}_{k+1} = (1-\lambda) \bar{\sigma}_k + \lambda \sigma_t$, where the superscripts represent the time steps and $\mu_t$, $\sigma_t$ denote the statistics estimated by the current test batch. 
GpreBN \cite{yang2022test} further combines the moving-averaged target statistics with the moving-averaged source statistics obtained during training to stabilize the estimation.
For stable and fast convergence, Mirza \etal~\cite{mirza2022norm} propose an adaptive moving average of normalization statistics that calculates specific statistics for each target sample. 
MemBN \cite{kang2024membn} introduces statistics
memory queues to store the test batch statistics and accumulate the latest test batch.
Combining the source and target statistics makes a trade-off between the target adaptation and the preservation of source discriminative ability. It also relaxes the requirements of the amount of target data. However, the hyperparameters for the combination need to be set according to different cases, which is inconvenient and may hurt robustness in real-world applications.

\vspace{.5mm}
\noindent
\textbf{Statistics inference.}
To achieve stable adaptation at test time for small amounts of target samples and varying distributions, MetaNorm \cite{du2020metanorm} infers the normalization statistics per sample.
Since it is difficult to estimate distribution statistics from a single sample, MetaNorm learns the ability under the meta-learning framework and supervises the inferred statistics by the domain-specific statistics during meta-training.
DCN \cite{jiang2023domain} also meta-learns the ability to infer target statistics from a single sample. Instead of directly inferring the normalization statistics, the method learns to estimate the channel-wise combination weights to combine the instance statistics of each target sample and the source statistics. 
Statistic inference avoids the requirement of larger amounts of data in normalization adaptation methods.
However, the inference is difficult to learn, which usually requires complex training strategies, which may lead to inefficient and unstable training.

\vspace{.5mm}
\noindent
\textbf{Discussion.}
Normalization adaptation methods achieve efficient adaptation by adjusting normalization statistics while fixing model parameters.
However, one major disadvantage is that these methods are not model-agnostic. Models without batch normalization layers, e.g., Vision Transformers \cite{dosovitskiy2020image}, will not benefit from these methods. Moreover, due to the inherent properties of batch normalization, these methods usually require large amounts of target samples for stable and robust adaptation.
Although there are various ways to alleviate this problem, e.g., instance normalization statistics, combination with source statistics, and statistics inference, they all have their own disadvantages.
The instance normalization statistics can result in less discrimination of the model. 
The combination of source and target statistics requires careful selection or training of the hyperparameters and update process to trade off the source and target information.
The statistics are difficult to estimate by a single sample, which requires complex methods and various source distributions to mimic the adaptation procedure during training.



\begin{figure*}[t]
\centering
\includegraphics[width=0.9\linewidth]{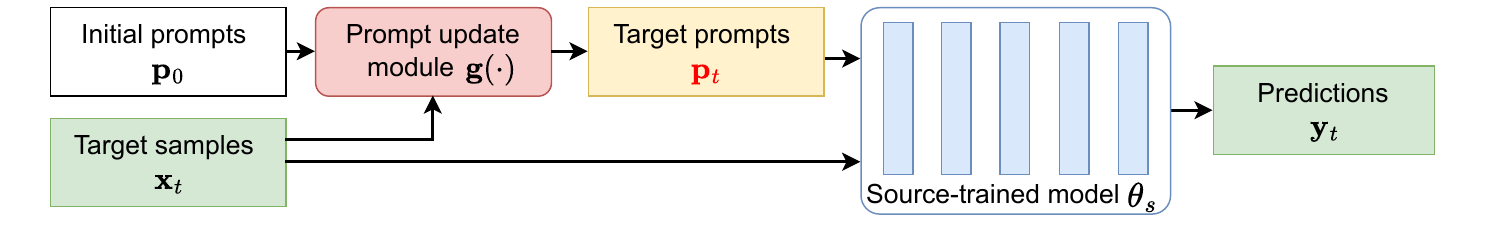}
\vspace{-2mm}
\caption{
\textbf{Prompt adaptation.} These methods update prompts for test samples, avoiding changing the model parameters.
}
\label{fig:promptada}
\vspace{-3mm}
\end{figure*}

\subsection{Sample adaptation}
As aforementioned, adapting model parameters can be computationally expensive and unstable, especially for large models, and normalization adaptation methods are limited to specific architectures~\cite{he2016deep}.
To solve these problems, sample adaptation methods are proposed to maintain the knowledge in the pretrained models at test time and instead adjust the target samples.
As illustrated in Figure \ref{fig4:sampleada}, sample adaptation methods adapt the target data $\x_t$ through module $\bpsi$, and make predictions on the adapted data $\x_s'$ by the fixed source-trained model $\btheta_s$:
\begin{equation}
\label{eq:samada}
    \textcolor{red}{\x_s'} = \bpsi(\x_t), ~~~ \y_t = f_{\btheta_s} (\textcolor{red}{\x_s'}).
\end{equation}
where $\bpsi$ denotes the adaptation module. 

Sample adaptation methods \cite{gao2022back, xiao2023energy, pandey2021generalization, nie2022diffusion} usually adapt their target data to the source data distributions through generative models, e.g., generative adversarial networks \cite{goodfellow2014generative}, variational auto-encoder \cite{kingma2013auto}, energy-based models \cite{hinton2002training, du2019implicit}, and diffusion models \cite{sohl2015deep, ho2020denoising}.
The generation-based methods are inspired by human visual recognition, where the challenging images are related to known familiar ones through iterative feedback for improved recognition \cite{huang2020neural, pandey2021generalization}.
Based on the generative models, these methods generate the corresponding source samples or features conditioned on the target ones and make predictions on the adapted data.
In these methods, the adaptation module $\bpsi$ in eq. \ref{eq:samada} denotes a trainable generative model, which is learned to model the source data distributions and adapt data from unseen distributions to the source distributions. Since the adaptation of the target samples can be conducted in both feature or input space, we categorize recent sample adaptation approaches into feature adjustment and input adjustment methods.

\vspace{.5mm}
\noindent
\textbf{Feature adjustment.}
Since features are learned to extract useful information from the input while eliminating redundant details, modeling and adjusting information in the feature space are efficient and easy to accomplish.
Thus, most sample adaptation methods are applied in the feature space.
Huang \etal~\cite{huang2020neural} utilize recurrent feedback for the out-of-distribution images. 
Their method adds generative feedback connections with latent variables to the neural networks.
The generative feedback updates the features of the target samples with the predictions and latent variables iteratively.
Pandey \etal~\cite{pandey2021generalization} first learn a domain invariant feature space from the source data, then train as generative model either a generative adversarial networks or a variational auto-encoder to generate features from the learned feature space. At test time, the method projects each target input to the source space by generating source features using the generative model, which is achieved by minimizing the distance between the generated and original target features.
Xiao \etal~\cite{xiao2023energy} train an energy-based model to update the target features to the source feature distribution. 
To achieve label-preserving sample adaptation, they further introduce a categorical latent variable that guides the update procedure.
Park \etal~\cite{park2023test} propose test-time style-shifting to transfer the target features to the nearest source distribution by the feature statistics.
A-star \cite{agarwal2023star} introduces attention segregation and retention loss functions to update the latent feature of the diffusion model for text-to-image generation.

\vspace{.5mm}
\noindent \textbf{Input adjustment.}
Building on the advances in generative models, recent methods have introduced sample adaptation in the input space \cite{gao2022back, tsai2024test}. Gao \etal~\cite{gao2022back} employ a diffusion model to directly transform the target images to the source images. The method first adds noise to the target image and iteratively updates the noisy input conditioned on the original images.
GDA \cite{tsai2024gda} further introduces structural guidance into the diffusion model for effectiveness and efficiency.
Oh \etal~\cite{oh2024efficient} fine-tunes an image editing model based on a latent diffusion model for sample adaptation to reduce resource requirements.

\vspace{.5mm}
\noindent \textbf{Discussion.}
Once the generative model is trained, sample adaptation methods achieve adaptation on each target sample without any extra data or fine-tuning operations. Therefore, these methods will be unaffected by the number of available target samples and are more stable in the case of varying target distributions. However, since the methods require iterative updates of the target samples, the efficiency will be worse than other methods. As comparison, once the fine-tuning is done on the target distribution, model adaptation methods only require a feedforward pass of the target samples, while inference adaptation and normalization adaptation methods do not even need fine-tuning operations.
These methods will be more efficient when the target distribution is already defined and sufficient data are available.

\subsection{Prompt adaptation}

Along with the development of hardware and computational resources, the amount of model parameters is becoming larger and larger.
The recent advances of foundation models in various tasks demonstrate the incredible ability of these large models \cite{achiam2023gpt, brown2020language, radford2021learning}.
However, because of the large amounts of parameters and training data, it is more and more difficult to adapt these large models, especially at test time with limited data.
Therefore, to efficiently adapt large models on downstream tasks, recent methods proposed prompt learning \cite{deng2023prompt, zhou2022conditional, zhou2022learning} and lightweight adapters \cite{gao2024clip, hu2021lora, wen2023batched}.
The prompting methods are further introduced into test-time adaptation, where the test data is more limited and dynamic. 
In test-time prompt adaptation, the prompt is first learned on the target data by:
\begin{equation}
    \textcolor{red}{\mathbf{p}_t} = \mathbf{g}(\x_t, \mathbf{p}_0),
\end{equation}
where $\mathbf{p}_0$ denotes the original prompt and $\mathbf{g}$ represents the learning method for prompt updating in a label-free manner, which can be achieved through gradient backpropagation, similar to model adaptation, or by generation via large language models. 
The learned prompt is then utilized to adapt the final predictions by:
\begin{equation}
    \mathbf{y}_t = f_{\btheta} (\x_t, \textcolor{red}{\mathbf{p}_t}),
\end{equation}
where $\btheta$ denotes the pretrained model parameters.
According to the different types of prompts $\mathbf{p}_t$, we categorize current prompt adaptation approaches into text-prompting and embedding-prompting methods.

%

\vspace{.5mm}
\noindent
\textbf{Text-prompting.}
Driven by advances in large language models, a common method for prompt adaptation is introducing appropriate textual prompts for specific tasks.
Text-based prompt adaptation is widely utilized in vision-language tasks to generate task-specific functions through updated textual features.
Some zero-shot learning methods \cite{hou2024domainverse, maniparambil2023enhancing, menon2022visual, roth2023waffling, liu2024zero} directly improve the text prompt by introducing corresponding descriptions or high-level concepts of each test class, which benefit from large language models like GPT-3 \cite{brown2020language} and GPT-4 \cite{achiam2023gpt}. 
PODA \cite{fahes2023poda} leverages CLIP to generate descriptions of the target domains for semantic segmentation.
Ren \etal~\cite{ren2024chatgpt} group the GPT-generated descriptions and construct knowledge trees for hierarchical classification.
To generate better text prompts, Mirza \etal~\cite{mirza2024meta} introduce meta-prompting to construct test-specific prompts by large language models under a hierarchical framework.
TEMPERA \cite{zhang2022tempera} gradually adapts the textual prompt by reinforcement learning.
OPT2I \cite{manas2024improving} utilizes large language models to iteratively update the prompt for each test sample for the text-to-image generation task. Beyond adapting prompts in the text space, there are also more general prompt adaptation methods conducted in the embedding space.

\vspace{.5mm}
\noindent
\textbf{Embedding-prompting.}
These methods leverage the advancements of the transformer architecture \cite{vaswani2017attention}, where the prompts are fed into the network together with the target input data to adjust the prediction function.
Some prompt adaptation methods \cite{gao2022visual, sun2023vpa} are based on the Vision Transformer \cite{jia2022visual}. Gao \etal~\cite{gao2022visual} tune the visual prompts by a hierarchical self-supervised loss function.
CVP \cite{tsai2024convolutional} designs convolutional visual prompts to use the convolutional structure as an inductive bias for visual test-time adaptation tasks.
U-VPA \cite{gan2023cloud} introduces an uncertainty-guided updating strategy for adaptation to continually changing environments.
Gan \etal~\cite{gan2023decorate} leverage visual prompt adaptation into continual test-time adaptation to avoid catastrophic forgetting.
TPGaze \cite{liu2024test} utilizes prompts to replace the original zero padding of the input images.
Niu \etal~\cite{niu2024test} propose forward-only prompt adaptation by a derivative-free optimizer, covariance matrix adaptation, to adjust prompts at test time.
He \etal~\cite{he2024domain} introduce domain
shift tokens at transformer layers and learn them by bilevel optimization in an adversarial manner.

There are also methods adapting prompts in the textual embedding space for multimodal models \cite{shu2022test, samadh2023align, xiao2024any, zhang2024robust, zhang2024historical}. 
TPT \cite{shu2022test} proposes prompt tuning for the multimodal CLIP model \cite{radford2021learning}.
To improve the effectiveness of prompt adaptation with limited data, DiffTPT \cite{feng2023diverse} utilizes pretrained diffusion models to generate diverse data for adaptation.
SwapPrompt \cite{ma2024swapprompt} leverages self-supervised contrastive learning to facilitate prompt adaptation.
Samadh \etal~\cite{samadh2023align} align the test sample statistics to the offline source statistics at test time.
C-TPT \cite{yoon2024c} explores calibration during prompt adaptation and considers the calibration error for optimizing the prompts.
DART \cite{liu2024dart} introduces both class-specific text prompts and instance-level image prompts, which are adaptively updated on subsequent test samples.
RLCF \cite{zhao2024test} further adopts the CLIP model as a reward model to provide feedback during test-time prompt tuning.
Unlike the optimization-based methods, any-shift prompting~\cite{xiao2024any} generates the test-specific prompt in a feedforward pass, without fine-tuning at test time.
Zanella \etal~\cite{zanella2024test} introduce test-time augmentation techniques to replace test-time prompt tuning for the CLIP model.
ZERO \cite{farina2024frustratingly} improves test-time prompt tuning methods by simply aggregating predictions on selected augmentations with zero softmax temperature.

\vspace{.5mm}
\noindent \textbf{Discussion.} 
Most prompt adaptation methods in the embedding space follow a model adaptation strategy by fine-tuning the prompts during test time. Unlike model adaptation, prompt adaptation does not require adjusting the model parameters, making it more computationally efficient, while preserving the model's generalization ability. The approaches are therefore more commonly used for the adaptation per target sample. Moreover, with the advancements in large language models, prompt adaptation is increasingly applied in the text space (e.g., GPT-generated descriptions), offering greater efficiency and explainability.



%% file: 3_methods_how.tex
 \section{How to prepare for adaptation}
\label{method:h2p}

Test-time adaptation methods tend to focus on the target data in the test stage, paying less attention to the preparation during source training. 
However, to improve the effectiveness and robustness of test-time adaptation, many methods also design specific training strategies. In this section, we categorize test-time adaptation methods according to their preparations during training as preparation-agnostic, training-preparation, and training-and-data preparation.

\vspace{.5mm}
\noindent\textbf{Preparation-agnostic.}
A prominent method for test-time model adaptation operates under the assumption that solely the source-trained model is available and all procedures are performed during the testing phase \cite{wang2021tent}.
Therefore, these methods are agnostic to the preparation in source training.

The preparation-agnostic model adaptation methods do not restrict or alter model training and only introduce unsupervised loss functions for fine-tuning at test time.
As introduced in Section \ref{sec:model-adapt}, model adaptation with entropy minimization \cite{wang2021tent, zhou2021bayesian, tang2023neuro, fleuret2021test}, pseudo-labeling \cite{rusak2021if, wang2022continual, chen2022contrastive, jang2022test, wang2023feature, boudiaf2022parameter}, and feature alignment \cite{wang2023feature, su2022revisiting, su2023revisiting, jung2022cafa, fleuret2021test, fang2013video} are the most widely-used methods for preparation-agnostic test-time adaptation since the loss functions are highly related to the main classification task.

Besides the model adaptation methods, most of the normalization adaptation methods \cite{schneider2020improving, nado2020evaluating, gong2022note, khurana2021sita, lim2023ttn} are also preparation-agnostic since they obtain the target normalization statistics by target batches and source statistics non-parametrically at test time.
Some inference adaptation methods, e.g., T3A \cite{iwasawa2021test} and AdaNPC \cite{zhang2023adanpc}, achieve parameter inference in a non-parametric manner at test time, which are also agnostic to the preparation during source training. Since altering the training stage of large models is data-intense and compute-heavy, most prompt adaptation methods \cite{shu2022test, fahes2023poda, zhang2022tempera, ma2024swapprompt, samadh2023align, zhao2024test} also tend to be preparation-agnostic.

\vspace{.5mm}
\noindent\textbf{Training preparation.}
In addition to the preparation-agnostic methods, some test-time adaptation approaches introduce additional source-training strategies or parameters to assist adaptation, which we categorize as methods with training preparation.

To improve the relation between self-supervised tasks and the main tasks, some model adaptation methods alter the source-training stage with extra self-supervision.
Test-time training \cite{sun2020test} and subsequent methods \cite{liu2021ttt++, gandelsmantest2022test, li2021test, osowiechi2023tttflow, cohen2020self} train the model jointly with the main task objective and the auxiliary objectives to improve the cooperation between main and auxiliary tasks.

Most inference adaptation methods also need extra preparation during source training since they require training extra modules for the inference of model parameters at test time \cite{dubey2021adaptive, kim2022variational, yeo2023rapid, wang2023dynamically}.
Although only estimating normalization statistics, the normalization adaptation methods InstCal-U \cite{zou2022learning} and TTN \cite{lim2023ttn} also require preparation since the method learns the parameters for combining source and target statistics in a post-training stage.

Sample adaptation methods \cite{gao2022back, pandey2021generalization, nie2022diffusion, huang2020neural} usually require preparations in training to model the source distribution. In addition to the neural network for the main task, these methods also need to train the generation model to encode the source distribution and learn to update the target data to the source at test time.

\input{tables/trainingset2}

\vspace{.5mm}
\noindent\textbf{Training-and-data preparation.}
In addition to altering training strategies, some methods \cite{alet2021tailoring, bartler2022mt3, du2020metanorm, sun2022dynamic, xiao2022learning} further change the training data to simulate distribution shifts during training to achieve more robust and generalizable test-time adaptation.
These methods require preparations for both the training strategy and the data in the source-training stage.

In model adaptation, training-and-data preparation is common for meta-learning-based methods
%
%
to strengthen the relationship between the auxiliary tasks and the primary task \cite{alet2021tailoring, bartler2022mt3, chen2022ost, chi2021test, cho2023complementary, choi2021test, hatem2023point, huangtest, liu2023meta, liu2022towards, min2023meta, sain2022sketch3t, zhong2022meta, wu2024test, sun2024learning}.
These methods follow the algorithm of model agnostic meta-learning \cite{finn2017model}, designing an inner loop to update the model of the auxiliary tasks and an outer loop to minimize the objectives of the primary task for the updated model parameters with specific data splits.
By doing so, the model learns good initialization parameters that can quickly adapt to unseen distributions with the auxiliary task \cite{sain2022sketch3t, huangtest}.
In other words, the model learns the underlying relationship between the primary and auxiliary tasks,
which benefits the adaptation with only the auxiliary tasks at test time.

Some inference adaptation methods, e.g., \cite{xiao2022learning, sun2022dynamic}, simulate domain shifts during training to learn the inference model that estimates domain-specific parameters with target samples.
The normalization adaptation methods \cite{du2020metanorm, zhang2021adaptive, xu2022mimic} mimic the target distribution to learn the ability to estimate statistics of unseen target distributions with small numbers of target samples.
To better adapt target samples to source distributions, some sample adaptation methods also require preparations on both training strategy and data. 
EBMDG \cite{xiao2023energy} utilizes different source data as positive and negative samples to train an energy-based model. Test-Time Style Shifting \cite{park2023test} introduces style balancing to handle domain-specific imbalance during source training. As a prompt adaptation method, Xiao \etal~\cite{xiao2024any} mimic distribution shifts during training to learn the ability to generate task-specific prompts at test time.

\vspace{.5mm}
\noindent
\textbf{Discussions.}
Methods that adapt only at test time, without altering the training stage \cite{wang2021tent, zhang2021memo, schneider2020improving} are more convenient for applications where the source data is hard to access or the source pretrained model is difficult to change, such as large foundation models.
Training-dependent adaptation methods design auxiliary tasks \cite{sun2020test, du2020metanorm, liu2021ttt++} or architectures \cite{xiao2023energy, pandey2021generalization, dubey2021adaptive, gao2022back} to assist the adaptation at test time, which is effective with extra knowledge but comes with a loss of efficiency during training.
Training-and-data preparation further enhances adaptation by simulating distribution shifts and learning the adaptation ability during training. However, these methods also bring more computational costs during training. To simulate distribution shifts, the methods are usually conducted on training data with multiple source distributions, which is also a limitation. We summarize representative test-time adaptation methods according to their training preparation strategy in Table \ref{trainingset}.




\section{How to adapt}
\label{method:h2a}

Once prepared, the adaptation settings are also important for the deployment of test-time adaptation algorithms in real-world applications. 
In this section, 
we categorize the existing methods according to their update strategies and inference data. 

%
%

\subsection{Update strategies}
According to their update strategy, we divide existing test-time adaptation methods into two categories, iterative update and on-the-fly update, which are highly related to computational costs and adaptation efficiency.
The iterative update methods usually require iterative optimizations of the model parameters or target samples to achieve adaptation, e.g., \cite{sun2020test, gao2022back}, whereas the on-the-fly update methods achieve adaptation in only a single pass of the data, e.g., \cite{xiao2022learning, du2020metanorm}.

\vspace{.5mm}
\noindent
\textbf{Iterative update.}
Most model adaptation methods \cite{sun2020test, wang2021tent, zhang2021memo, wang2022continual, niu2023towards, chen2022contrastive, goyal2022test, jang2022test} are iterative at inference since they require step-by-step fine-tuning to update their model parameters at test time \cite{sun2020test, liu2021ttt++, gandelsmantest2022test, wang2021tent, shu2022test, lee2022surgical}. 
Since fine-tuning usually comes with high computational costs, some model adaptation methods propose fine-tuning only a subset of the model parameters at test time to improve their adaptation efficiency and stability. Tent \cite{wang2021tent} only updates the statistics and affine parameters of the batch normalization layers, which is followed by many \cite{sinha2023test, zhang2021memo, gordon2018meta, yuan2024tea}. Others introduce additional lightweight adaptive models and only update the adaptive parameters for efficient inference \cite{chen2023improved, song2023ecotta}. 
Recent works find that different models \cite{lisimple} or fine-tuning different blocks \cite{lee2022surgical} performs best for different distribution shifts.
Therefore, Lee \etal~\cite{lee2022surgical} propose to selectively adapt a subset of the model parameters according to the distribution shifts.
Tang \etal~\cite{tang2023neuro} propose to learn lower layer representations by a feedforward Hebbian learning layer.
By adapting the model parameters partially, these methods achieve a good trade-off between computational costs and adaptation performance.

Like model adaptation, most sample adaptation methods \cite{gao2022back, xiao2023energy, pandey2021generalization, huang2020neural} also require iterative updates by updating the target samples to the source distributions step-by-step through generative models like energy-based model and diffusion model.
Many prompt adaptation methods also follow the adaptation strategy of model adaptation, therefore also requiring iterative inference at test time \cite{shu2022test, ma2024swapprompt, yoon2024c, gao2022visual, sun2023vpa, fahes2023poda, samadh2023align, zhang2022tempera, manas2024improving, zhao2024test}.
Although avoiding fine-tuning the model parameters, they adapt the prompts by iteratively updating them by backpropagation \cite{shu2022test, ma2024swapprompt, samadh2023align} or large language model feedback \cite{zhao2024test, zhang2022tempera, manas2024improving} at test time.

\vspace{.5mm}
\noindent
\textbf{On-the-fly update.}
Different from the iterative methods, on-the-fly update achieves adaptation in a single feedforward pass at test time, without the need for fine-tuning or backpropagation.
All inference adaptation methods (e.g., \cite{iwasawa2021test, zhang2023adanpc, huangextrapolative, dubey2021adaptive, xiao2022learning, yeo2023rapid, sun2022dynamic, kim2022variational}) perform an on-the-fly update since these methods directly infer the model parameters at test time.
The normalization adaptation methods \cite{du2020metanorm, schneider2020improving, lim2023ttn, gong2022note, nado2020evaluating} are also on the fly, by estimating their normalization statistics in a single feedforward pass.
 vAs a sample adaptation method, test-time style shifting \cite{park2023test} achieves an on-the-fly update by adapting the target features through the distribution statistics, without iterative update.
Some prompt adaptation methods can also be on-the-fly at inference by utilizing pretrained large models \cite{fahes2023poda, roth2023waffling, ren2024chatgpt, mirza2024meta} or by learning the ability of prompt generation during training \cite{xiao2024any}.

\vspace{.5mm}
\noindent
\textbf{Discussions.}
Given enough time and computational resources, iterative update methods can yield reliable adaptation results. However, they may struggle in real-world applications that have strict computational cost requirements. On-the-fly update is more time-efficient for adaptation and prediction during test time, making them better suited for real-time applications. Nevertheless, these methods can be constrained by complex training strategies \cite{xiao2022learning, yeo2023rapid, du2020metanorm} or specific applications \cite{iwasawa2021test, schneider2020improving, lim2023ttn}.

\subsection{Inference data}
\label{sec: h2a-data}
Besides the update strategy, the accessible target data is also an essential factor for test time adaptation.
In real-world applications, target distributions are typically unknown, making it challenging to collect sufficient data from specific target distributions. 
Therefore, the inference data is crucial for the deployment of test-time adaptation approaches.
Based on the data requirements of the target distribution, we classify current test-time adaptation methods into four categories: online, batch-wise, sample-wise, and dynamic inference.

\vspace{.5mm}
\noindent
\textbf{Online inference.}
With the goal of achieving adaptation along with inference, online inference is a popular strategy in test-time adaptation to handle the online target data and reuse the target information from previous target samples. 

Online inference is widely utilized in model adaptation methods. Following test-time training \cite{sun2020test} and Tent \cite{wang2021tent}, these methods achieve inference by initializing the model parameters of each target batch with the parameters adapted by previous target batches \cite{liu2021ttt++, zhou2021bayesian, goyal2022test, chen2022contrastive, lee2022surgical}.
Some inference adaptation methods \cite{iwasawa2021test, zhang2023adanpc, huangextrapolative, alfarra2024evaluation} also achieve online inference. T3A \cite{iwasawa2021test} adjusts a prototype-based classifier online with target mini-batches. 
AdaNPC \cite{zhang2023adanpc} updates a target-specific memory online and makes predictions based on the updated memory. Some normalization adaptation methods \cite{yang2022test, mirza2022norm, hong2023mecta, zhao2023delta} continually update the normalization statistics at test time to generate more representative and reliable target statistics.

Despite their widespread use and significant advances, online inference methods typically assume that the online target data originates from the same target distribution. Consequently, these methods struggle with complex settings, such as having limited target data and intricate target distributions, where the methods may lead to error accumulation or knowledge forgetting.

\vspace{.5mm}
\noindent\textbf{Batch-wise inference.} 
Batch-wise inference methods achieve adaptation for each batch of target samples to avoid error accumulation among different batches.
Normalization adaptation methods \cite{schneider2020improving, zhang2021adaptive, nado2020evaluating, lim2023ttn, you2021test} usually focus on batch-wise inference since target batch statistics are important for the statistics estimation. Some inference adaptation methods \cite{dubey2021adaptive, yeo2023rapid} also utilize the information within each target batch to infer target-specific model parameters. Batch-wise inference reduces error accumulation in online settings. However, it still requires each batch of samples to originate from the same target distribution, which may not be true in real-world applications.

\vspace{.5mm}
\noindent\textbf{Sample-wise inference.}
To further reduce the requirement of target sample amounts and fit complex target distributions, sample-wise inference methods are proposed to achieve adaptation for each individual target sample \cite{cohen2020self, d2019learning, hong2021deep, he2020self, gao2022back, jiang2022test, shu2022test, zhang2022auxadapt}.
By adapting each target sample, the methods can handle test data from various target distributions without error accumulation and forgetting.
Through updating each target sample individually, all sample adaptation methods \cite{gao2022back, xiao2023energy, pandey2021generalization, huang2020neural, park2023test} belong to the sample-wise regime.
Some model adaptation methods achieve sample-wise inference by fine-tuning their model parameters on the augmentations of each target sample \cite{sun2020test, zhang2021memo}.
SiSTA \cite{thopalli2023target} fine-tunes a generative model using a single-shot target, and samples the synthetic target data for adaptation at test time.
ViTTA \cite{lin2022video} deploys temporal augmentation for sample-wise test-time adaptation on video data. 
Meta-learning is also utilized in model adaptation methods to learn the ability of sample-wise inference during training \cite{alet2021tailoring, bartler2022mt3, sain2022sketch3t, min2023meta}.
Similarly, some normalization adaptation methods estimate representative normalization statistics in a sample-wise fashion also by data augmentation \cite{khurana2021sita, mirza2022norm, hu2021mixnorm} or meta-learning \cite{du2020metanorm, jiang2023domain}.
Replacing batch normalization with instance normalization is also an easy and efficient method for sample-wise test-time adaptation \cite{kaku2020like, gong2022note}.
Some inference adaptation methods \cite{xiao2022learning, wang2023dynamically} also meta-learn the inference module to infer target-specific model parameters through a single target sample.
Following the model adaptation methods \cite{zhang2021memo}, a large subset of prompt adaptation methods \cite{shu2022test, ma2024swapprompt, samadh2023align, yoon2024c} achieves adaptation per sample based on data augmentation strategies.

Sample-wise inference methods achieve test-time adaptation without the requirement of having access to large amounts of data from the same distribution, avoiding error accumulation and forgetting problems. However, these methods can not benefit from more target data, which limits their adaptation performance.

\input{tables/adaptset2}

\vspace{.5mm}
\noindent
\textbf{Dynamic inference.}
To handle catastrophic forgetting in continually changing environments with various distributions, methods for test-time adaptation in dynamic scenarios are proposed to achieve more stable and reliable online adaptation on complex target distributions \cite{brahma2022probabilistic, chakrabarty2023sata, dobler2023robust, hong2023mecta, liu2023vida, niloy2024effective, niu2022efficient, niu2023towards, song2023ecotta, wang2022continual, yuan2023robust, press2023rdumb, gui2024active, yang2024versatile, liu2024continual, park2024ul, jiang2024pcotta}.
Many model adaptation methods rely on dynamic inference. To deal with changing environments at test time, Continual TTA \cite{wang2022continual} introduces continual model adaptation and preserves the knowledge obtained during training by stochastically restoring a small part of the source-trained weights in each iteration.
Niu \etal~\cite{niu2022efficient} proposes an anti-forgetting regularizer with the Fisher importance \cite{kirkpatrick2017overcoming} of model weights.
SAR \cite{niu2023towards} replaces batch normalization with layer normalization \cite{ba2016layer} and group normalization \cite{wu2018group} for efficient test-time adaptation on small batch sizes.
Brahma \etal~\cite{brahma2022probabilistic} utilizes Fisher-information-based model restoration with Bayesian adaptation \cite{zhou2021bayesian} for continual model adaptation.
EcoTTA \cite{song2023ecotta} introduces a self-distilled regularization on the outputs of the adapted and source-trained parameters to prevent forgetting. 
BECoTTA \cite{lee2024becotta} proposes Mixture of Domain Low-rank Experts to selectively capture knowledge for each test domain.
Zhu \etal~\cite{zhu2024reshaping} introduces an uncertainty-aware buffer to aggregate significant samples with high certainty and provide reliable signals for the adaptation process.
Hoang \etal~\cite{hoang2024persistent} proposes a recurring setting where environments not only change but also recur over time.
The sample-wise inference methods can also be utilized in dynamic scenarios \cite{niu2022efficient, niu2023towards} since they are unaffected by changing distributions.

\vspace{.5mm}
\noindent
\textbf{Discussions.}
Online inference is extensively researched and applied in various scenarios. Test-time adaptation methods relying on online inference can achieve reliable adaptation when sufficient computational resources and target data are available. However, they struggle in real-world applications lacking adequate data from the same target distribution, leading to potential issues like error accumulation and knowledge forgetting.
By contrast, batch-wise inference methods prevent error accumulation, but still require a batch of data from the same target distribution. Sample-wise inference goes further by adapting to each target sample individually, resulting in robust performance in complex applications without strict requirements for large amounts of target data. These methods are more effective in dealing with test data from unknown distributions. However, they cannot benefit progressively from additional target data, which limits their performance.
Dynamic inference addresses the adaptation of continually changing distributions without knowledge forgetting. However, these methods typically assume the availability of distribution annotations, reducing the flexibility for use in more complex scenarios. We summarize representative test-time adaptation methods according to their update strategy and inference data in Table \ref{adaptset}.



%% file: tables/trainingset2.tex
\begin{table*}[t!]
\begin{center}
\vspace{-2mm}
\caption{\textbf{How to prepare for adaptation.} This table summarizes representative test-time adaptation methods according to what they adapt (Section~\ref{method:w}) and their training preparation strategy (Section~\ref{method:h2p}).}
\label{trainingset}
\vspace{-2mm}
	\resizebox{1.8\columnwidth}{!}{%
		\setlength\tabcolsep{4pt} 
\begin{tabular}{lccc}
\toprule
         & Preparation-agnostic & Training preparation & Training \& Data preparation \\ 
        \midrule
        \rowcolor{mColor2}
        \multicolumn{4}{l}{\textbf{Model adaptation}} \\
        Auxiliary self-supervision & - & \cite{liu2021ttt++, gandelsmantest2022test, li2021test, osowiechi2023tttflow, cohen2020self, sun2020test} & \cite{alet2021tailoring, bartler2022mt3, chi2021test, choi2021test, hatem2023point, zhong2022meta} \\ 
        Entropy minimization & \cite{wang2021tent, zhou2021bayesian, tang2023neuro, zhang2021memo, niu2022efficient, lee2022surgical} & - & - \\
        Pseudo-labeling & \cite{rusak2021if, wang2022continual, chen2022contrastive, jang2022test, goyal2022test} & - & \cite{ambekarvariational} \\
        Feature alignment & \cite{wang2023feature, su2022revisiting, jung2022cafa, fleuret2021test, mirza2022actmad} & - & - \\
        \midrule
        \rowcolor{mColor2}
        \multicolumn{4}{l}{\textbf{Inference adaptation}}  \\
        Batch-wise inference & \cite{iwasawa2021test, huangextrapolative, zhang2023adanpc} &
        \cite{dubey2021adaptive} & - \\ 
        Sample-wise inference & \cite{zhang2024boostadapter, karmanov2024efficient, zhang2024dual} &
        \cite{yeo2023rapid, kim2022variational, wang2023dynamically} & \cite{xiao2022learning, sun2022dynamic} \\ 
        \midrule
        \rowcolor{mColor2}
        \multicolumn{4}{l}{\textbf{Normalization adaptation}} \\
        Target statistics & \cite{gong2022note, kaku2020like, li2016revisiting, nado2020evaluating}
         & -  &  \cite{zhang2021adaptive}  \\
        Statistics combination &
        \cite{schneider2020improving, khurana2021sita, you2021test, hu2021mixnorm, yang2022test} & \cite{lim2023ttn, zou2022learning}  &  -  \\
        Statistics inference & - & -  &  \cite{du2020metanorm, xu2022mimic}  \\
        \midrule
        \rowcolor{mColor2}
        \multicolumn{4}{l}{\textbf{Sample adaptation}} \\
        Feature adjustment & - &
        \cite{pandey2021generalization, nie2022diffusion, huang2020neural}   & \cite{xiao2023energy, park2023test} \\ 
        Input adjustment & - &
        \cite{gao2022back, oh2024efficient, tsai2024gda}   & - \\ 
        \midrule
        \rowcolor{mColor2}
        \multicolumn{4}{l}{\textbf{Prompt adaptation}} \\
        {Text-prompting} & \cite{roth2023waffling, ren2024chatgpt, fahes2023poda, zhang2022tempera} & - & - \\ 
        {Embedding-prompting} & \cite{shu2022test, yoon2024c, gao2022visual, ma2024swapprompt, zhao2024test} & - & \cite{xiao2024any} \\  
\bottomrule
\end{tabular}
}
\vspace{-2mm}
\end{center}
\end{table*}


%% file: tables/adaptset2.tex
\begin{table*}[t]
\begin{center}
\vspace{-2mm}
\caption{\textbf{How to adapt.} This table summarizes representative test-time adaptation methods according to what they adapt (Section~\ref{method:w}) and their update strategy and inference data (Section~\ref{method:h2a}).
}
\label{adaptset}
	\resizebox{2\columnwidth}{!}{%
		\setlength\tabcolsep{4pt} 
\begin{tabular}{lcccccc}
\toprule
         & \multicolumn{2}{c}{Update strategy} & \multicolumn{4}{c}{Inference data} \\
        \cmidrule(lr){2-3} \cmidrule(lr){4-7}
        & Iterative & On-the-fly & Online & Batch-wise & Sample-wise & Dynamic \\ 
        \midrule
        \rowcolor{mColor2}
        \multicolumn{7}{l}{\textbf{Model adaptation}} \\
        Auxiliary self-supervision  & \cite{sun2020test, liu2021ttt++, min2023meta, gandelsmantest2022test} & - & \cite{sun2020test, liu2021ttt++, chen2023improved} & - & \cite{bartler2022mt3, alet2021tailoring, sun2020test} & \cite{yuan2023robust, dobler2023robust} \\ 
        Entropy minimization & \cite{wang2021tent, niu2022efficient, lee2022surgical, zhang2021memo} & - & \cite{wang2021tent, zhou2021bayesian, lee2022surgical} & - & \cite{zhang2021memo} & \cite{niu2022efficient, hong2023mecta} \\
        Pseudo-labeling & \cite{wang2022continual, jang2022test, goyal2022test, chen2022contrastive} & - & \cite{jang2022test, goyal2022test, chen2022contrastive} & - & - & \cite{song2023ecotta, wang2022continual} \\
        Feature alignment & \cite{wang2023feature, su2022revisiting} & - & - & - & - & - \\
        \midrule
        \rowcolor{mColor2}
        \multicolumn{7}{l}{\textbf{Inference adaptation}} \\
        Batch-wise inference & - & \cite{iwasawa2021test, zhang2023adanpc, dubey2021adaptive} & \cite{iwasawa2021test, huangextrapolative, zhang2023adanpc} & \cite{dubey2021adaptive} & - & - \\
        Sample-wise inference & - & \cite{xiao2022learning, yeo2023rapid} & - & - & \cite{xiao2022learning, wang2023dynamically, sun2022dynamic} & - \\ 
        \midrule
        \rowcolor{mColor2}
        \multicolumn{7}{l}{\textbf{Normalization adaptation}} \\
        Target statistics & - & \cite{gong2022note, nado2020evaluating} & \cite{yang2022test, mirza2022norm} & \cite{zhang2021adaptive, you2021test} & \cite{kaku2020like, gong2022note, tomar2024mixing} & - \\
        Statistics combination & - & \cite{schneider2020improving, lim2023ttn} & \cite{hong2023mecta, zhao2023delta} & \cite{nado2020evaluating, lim2023ttn} & \cite{khurana2021sita, mirza2022norm, hu2021mixnorm} & - \\
        Statistics inference & - & \cite{du2020metanorm} & - & - &  \cite{du2020metanorm, jiang2023domain} & - \\ 
        \midrule
        \rowcolor{mColor2}
        \multicolumn{7}{l}{\textbf{Sample adaptation}} \\
        Feature adjustment & \cite{gao2022back, pandey2021generalization, huang2020neural, xiao2023energy} & \cite{park2023test} & - & - & \cite{gao2022back, pandey2021generalization, huang2020neural, xiao2023energy, park2023test} & - \\ 
        Input adjustment & \cite{gao2022back} & - & - & - & \cite{gao2022back} & - \\ 
        \midrule
        \rowcolor{mColor2}
        \multicolumn{7}{l}{\textbf{Prompt adaptation}} \\
        Text-prompting & \cite{zhang2022tempera, manas2024improving} & \cite{fahes2023poda, roth2023waffling, ren2024chatgpt} & - & - & \cite{ren2024chatgpt, fahes2023poda, manas2024improving} & - \\
        Embedding-prompting & \cite{shu2022test, yoon2024c, samadh2023align, zhao2024test} & \cite{xiao2024any, niu2024test, farina2024frustratingly} & \cite{niu2024test, liu2024dart} & - & \cite{shu2022test, ma2024swapprompt, samadh2023align, xiao2024any} & - \\
        \bottomrule
\end{tabular}
}
\vspace{-2mm}
\end{center}
\end{table*}

%% file: 4_application.tex


\input{tables/classification_setting}

\section{Evaluating test-time distribution shifts}
\label{app:classification}


Existing methods are predominantly evaluated on image classification tasks, as implementing various distribution shifts is relatively straightforward for these tasks. Consequently, image classification is the most extensively studied task in test-time adaptation, regardless of the specific adaptation techniques used \cite{sun2020test, wang2021tent, xiao2022learning, zhang2021memo, gao2022back, schneider2020improving, xiao2023energy, du2020metanorm, shu2022test, yoon2024c}. Evaluations on image classification tasks are straightforward. After training a model on one or more source distributions, adaptation and evaluation are conducted using the source-trained model and target samples.
There are various tasks and corresponding benchmarks for image classification, each with different types of distribution shifts. In this section, we will detail the evaluations for image classification under covariate shifts with single or multiple source distributions, as well as label shifts, conditional shifts, and joint shifts.

%

\vspace{.5mm}
\noindent
\textbf{Covariate shifts with single source distribution.}
Image classification across covariate shifts is the most widely applied evaluation task for test-time adaptation. One commonly investigated task is the single-source distribution problem, where the model is trained on a single source distribution and adapted to various target distributions at test time \cite{sun2020test, wang2021tent, zhao2023pitfalls}. Several benchmarks with various covariate shifts allow to evaluate adaptation approaches in single-source image classification, e.g., natural, corruptions, image styles, environments, etc.

\begin{figure}[t]
\centering
\vspace{-2mm}
\includegraphics[width=0.99\linewidth]{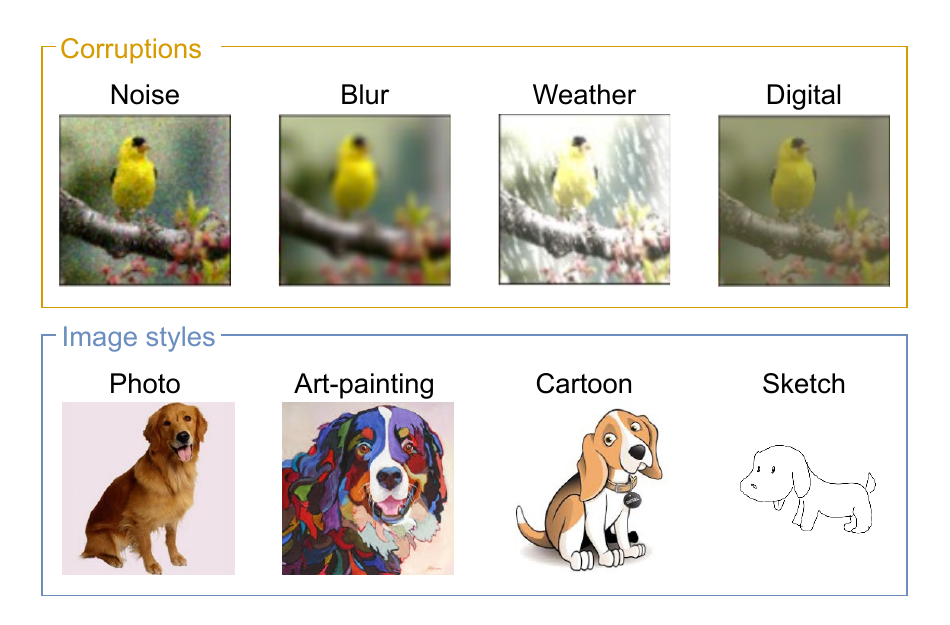}
\vspace{-2mm}
\caption{\textbf{Examples of covariate shifts in image classification.} Corruptions (first row) are usually evaluated for single-source settings, while image styles are usually evaluated in multi-source settings, following the common experimental setup in domain generalization.}
\label{corruption}
\vspace{-4mm}
\end{figure}

One common covariate shift utilized in test-time adaptation is the corruption \cite{hendrycks2019benchmarking}, e.g., CIFAR-10-C, CIFAR-100-C, and ImageNet-C \cite{sun2020test, wang2021tent, schneider2020improving, gao2022back, zhang2021memo, lim2023ttn}. A model is trained on the original dataset, e.g., CIFAR \cite{krizhevsky2009learning} or ImageNet \cite{russakovsky2015imagenet}.  
The target distributions are obtained by applying 15 types of corruption at five severity levels on the original test set of the datasets, some samples are shown in Figure \ref{corruption}.

Digit adaptation is a common benchmark in domain adaptation, where different distributions are different datasets, e.g., MNIST \cite{lecun1998gradient}, MNIST-M \cite{ganin2015unsupervised}, SVHN \cite{netzer2011reading}, and USPS \cite{hull1994database}. The distributions have the same label space with 10 categories. Following \cite{wang2021tent}, most test-time adaptation methods on digits datasets train their model on SVHN. The source-trained model is then adapted and evaluated on the different target distributions of MNIST, MNIST-M, and USPS.

There are also other test-time adaptation datasets based on common image classification benchmarks like CIFAR and ImageNet.
\texttt{CIFAR-10.1} \cite{recht2018cifar} is a test set for CIFAR-10 that consists of images not present in the original CIFAR-10 dataset.
While both datasets share the same 10 classes, the images in CIFAR-10.1 are selected to be more challenging and less similar to the CIFAR-10 training set, which is referred to as the natural distributional shift.
\texttt{ImageNet-A} \cite{hendrycks2021natural} also focuses on natural distribution shifts. It contains real-world, unmodified, and naturally occurring examples of ImageNet classes.
\texttt{ImageNet-(S)ketch} \cite{wang2019learning} consists of sketch-like images, which match the ImageNet classification validation set in categories and scale.
\texttt{ImageNet-R} \cite{hendrycks2021many} considers distribution shifts of various renditions of the original images, e.g., paintings, cartoons, etc.

The single-source image classification task has been explored using various kinds of test-time adaptation approaches, including model, inference, normalization, sample, and prompt adaptation.
The absence of sample-wise inference adaptation and statistics inference normalization methods in this setting can be attributed to the fact that these methods typically require simulating distribution shifts during training, which necessitates multiple source distributions.

\begin{tikzpicture}[remember picture, overlay]
\node at (13.3,11.8) [fill=yellow!50,opacity=0.3,text opacity=0.5,rounded corners=5pt, minimum width=8.1cm, minimum height=3.8cm, align=center] 
{{Emerging research opportunity I:} \\ \textbf{Beyond model adaptation and covariate shifts}};
\end{tikzpicture}

\noindent
\textbf{Covariate shifts with multiple source distributions.}
Test-time adaptation methods that leverage multiple source distributions during training may learn extra models for domain information extraction \cite{dubey2021adaptive, iwasawa2021test} or simulate distribution shifts during training \cite{du2020metanorm, xiao2022learning}.
In these methods, domain generalization datasets are utilized widely.
Since domain generalization has been investigated for a long time there exist various datasets, e.g., PACS, Office-Home, and DomainNet, which cover multiple distributions from different domains.
PACS \cite{li2017deeper} consists of 9,991 images of seven classes from four domains, i.e., \textit{photo}, \textit{art-painting}, \textit{cartoon}, and \textit{sketch}.
\textit{VLCS} \cite{fang2013video} consists of 5 classes from 4 different datasets: Pascal, LabelMe, Caltech, and SUN. 
Office-Home \cite{venkateswara2017deep} also contains four domains, i.e., \textit{art}, \textit{clipart}, \textit{product}, and \textit{real-world}, which have 15,500 images of 65 categories in total. 
\textit{TerraIncognita} \cite{beery2018recognition} has four domains taken by camera from four different locations. The dataset includes 24,778 samples of 10 categories.
DomainNet \cite{peng2019moment} is more challenging since it has six domains i.e., \textit{clipart}, \textit{infograph}, \textit{painting}, \textit{quickdraw}, \textit{real}, \textit{sketch}, with 586,575 examples of 345 classes.
The methods conducted on these domain generalization datasets follow the ``leave-one-out'' protocol \cite{li2017deeper}, where one domain is utilized as the target domain and the remaining domains act as the source domains. The source model is trained on all source domains and then adapted and evaluated on the target domain.
A benchmark of test-time adaptation methods on these datasets is provided by Yu \etal \cite{yu2023benchmarking} and Alfarra \etal \cite{alfarra2023revisiting}. 
Beyond these datasets with different image styles, WILDS \cite{koh2021wilds} contains 10 datasets across a more diverse set of application areas, data modalities, and dataset sizes. Each dataset comprises data from different domains.

Similar to single-source tasks, all test-time adaptation approaches can be applied to these multi-source image classification tasks. 

\vspace{.5mm}
\noindent
\textbf{Other covariate shift tasks.}
Based on image classification benchmarks, there are several emerging test-time adaptation scenarios, e.g., having continually changing distributions \cite{brahma2022probabilistic, song2023ecotta, wang2022continual, yuan2023robust, press2023rdumb} or learning with limited amounts of target data \cite{gao2022back, xiao2022learning, zhang2021memo}, which we discussed in Section \ref{sec: h2a-data}.


\vspace{.5mm}
\noindent
\textbf{Label shifts.}
In addition to covariate shifts, other types of distribution shifts are also investigated in image classification tasks. 
Label shifts are explored in tasks with spurious correlations \cite{sun2024beyond, lee2022surgical}, long-tailed data \cite{park2023label}, or zero-shot data \cite{xiao2024any, maniparambil2023enhancing}, where the distributions of the target labels differ from the source ones. 
Spurious correlation are utilized as label shifts by Sun \etal \cite{sun2024beyond}, where additional meta-data labels are introduced as group/attribute labels for the inputs. The changes in the relationships between real labels $y$ and the attribute labels are treated as label shifts. In other words, the real labels are spuriously correlated with the attributes. CelebA and Waterbirds are utilized to evaluate this type of label shifts \cite{sagawa2019distributionally, lee2022surgical, sun2024beyond}. The real labels and attribute labels are the hair color and sex respectively in CelebA while bird type and background type respectively in Waterbirds. In addition, Colored-MNIST and CheXpert are also utilized by Sun \etal ~\cite{sun2024beyond}.
Park \etal~\cite{park2023label} consider long-tailed data to achieve label shifts, where the distribution of real labels $p(y)$ differs between the training and test sets. In their approach, the training set is long-tailed using CIFAR-10/100-C \cite{cao2019learning} or ImageNet-C \cite{liu2019large}, while the test set follows the common setting with balanced data for each category.
Zero-shot learning can also be treated as a challenging version of label shifts, where $p(\y) {=} 0$ for the unknown classes during training. This setting is more common in prompt adaptation methods \cite{xiao2024any, maniparambil2023enhancing}, evaluated on different datasets such as ImageNet, UCF101, and Oxford-Flowers, following the base-to-new classification settings proposed by Zhou \etal ~\cite{zhou2022conditional}.

\vspace{.5mm}
\noindent
\textbf{Conditional shifts.}
Conditional shifts usually assume the distribution of labels is the same, while the relationship between samples and labels is different.
Sub-population datasets, such as Living-17 and Entity-30 \cite{santurkar2020breeds}, commonly evaluate methods across conditional shifts \cite{lee2022surgical, xiao2024any}, where the source and target distributions consist of the same classes but contain samples from different subclasses of those classes.

\vspace{.5mm}
\noindent
\textbf{Joint shifts.}
Test-time adaptation is also evaluated for joint distribution shifts, tackling covariate and label shifts simultaneously \cite{park2023label, xiao2024any}. Park \etal~\cite{park2023label} combine corruption and long-tailed data, training the model on long-tailed clean data while adapting it to corrupted data with balanced categories. Other methods evaluate performance across joint shifts in open-set classification. Gao~\etal ~\cite{gao2024unified} and Lee~\etal ~\cite{lee2023towards} pretrain their model on clean training sets and adapt it to corrupted data with unseen categories, such as SVHN-C for CIFAR-10/100 and ImageNet-O-C for TinyImageNet-C. The Office-Home dataset is also used for open-set settings to combine covariate shifts in image styles and label shifts in unseen categories \cite{xiao2024any}.


Overall, covariate shifts have been evaluated most in test-time adaptation, with various settings and different adaptation approaches.
In contrast, label, conditional, and joint shifts make changes at the input level and
introduce new knowledge at the semantic level. Therefore, most methods addressing these distribution shifts focus on model adaptation and prompt adaptation. The additional supervision in model adaptation and the extra knowledge from large pretrained models in prompt adaptation provide the necessary information to effectively address these distribution shifts. Other adaptation methods may also benefit from extra knowledge in solving these distribution shifts. We summarize representative test-time adaptation methods that evaluate various distribution shifts of image classification in Table \ref{tab: classapp}.

\section{Applications}
\label{application}


Since distribution shift commonly exists in various areas, test-time adaptation has many applications. 
In this section, we categorize the test-time adaptation methods based on 
the application of the methods to various tasks, including image-level, video-level, and 3D-level tasks, as well as tasks beyond vision applications.

\subsection{Image-level applications}

As one of the most extensively studied areas in machine learning, image-level tasks are prime candidates for test-time adaptation due to the common distribution shifts encountered in real-world scenarios. 

\vspace{.5mm}
\noindent \textbf{Image classification.}
Image classification is the most investigated task in test-time adaptation due to its ease of implementing various distribution shifts, straightforward evaluation, and generalization to other tasks. 
Test-time adaptation algorithms are usually first conducted on image classification and then extended to other tasks. We discussed the commonly used distribution shifts and settings for image classification in Section \ref{app:classification}.

\vspace{.5mm}
\noindent \textbf{Dense prediction.}
Dense prediction is one of the important image-level applications where predictions are made at the pixel level.
These applications are crucial for understanding the detailed structure and content of images, enabling precise and localized decisions, for example, image segmentation, detection, depth estimation, and optical flow estimation.
Since distribution shifts exist widely in images, test-time adaptation approaches are also investigated in the dense prediction tasks \cite{yeo2023rapid, bahmani2022semantic, park2024test, sinha2023test, yang2024exploring, jang2024talos}.

Image segmentation is the most popular among the image-level dense prediction tasks. Image segmentation may encounter distribution shifts in automatic driving scenarios, with continually changing surrounding environments.
The adaptation is then achieved for various unseen environments at test time \cite{bahmani2022semantic, colomer2023adapt, khurana2021sita, volpi2022road, wang2023dynamically, zou2022learning, lee2023towards, hu2024relax}.
The methods utilize one or several of the datasets as source distributions and the remaining ones as the target dataset to simulate distribution shifts between synthetic and real-world or different environments. Except for segmentation, test-time adaptation methods for depth estimation \cite{izquierdo2023sfm, park2024test, yeo2023rapid, zhao2024metric}, object detection \cite{sinha2023test, veksler2023test}, and dense correspondence \cite{hong2021deep} have also been proposed recently.

Most test-time adaptation methods for dense prediction are based on model adaptation \cite{izquierdo2023sfm, park2024test, hong2021deep, lee2023towards, sinha2023test, veksler2023test} since dense prediction results provide more information to update model parameters.
There are also inference adaptation \cite{yeo2023rapid, wang2023dynamically}, normalization adaptation \cite{khurana2021sita, zou2022learning}, and prompt adaptation methods for more efficient adaptation at test time. 
As of yet, sample adaptation is seldom used in dense prediction tasks because these tasks rely on detailed local information, which cannot be delivered by existing generative models during sample adaptation.

\vspace{.5mm}
\noindent \textbf{Image enhancement.}
Improving image quality is crucial for many practical applications, which may encounter distribution shifts. 
Consequently, test-time adaptation methods are proposed for image enhancement such as super-resolution \cite{deng2023efficient, rad2021test} and low-light image enhancement \cite{yao2023generalized}.
Additionally, to improve the quality of noisy images, test-time adaptation is applied in image restoration \cite{gou2024test}, denoising \cite{mansour2024tttmim}, dehazing \cite{liu2022towards, chen2024prompt}, and deblurring \cite{chi2021test}.
Moreover, there are also test-time adaptation efforts aimed at image quality assessment \cite{roy2023test}.
Image enhancement test-time adaptation methods are mainly based on model adaptation, as well as normalization adaptation. Since it is difficult to obtain entropy or pseudo-labels for these tasks, the methods usually leverage auxiliary self-supervision or consistency loss functions like reconstruction loss \cite{deng2023efficient, chi2021test, gou2024test}. Additionally, because these tasks are more abstract than classification, managing them with inference adaptation and prompt adaptation is challenging without task-specific descriptions or representations. Like dense prediction, image enhancement tasks require detailed information, which is difficult to preserve through sample adaptation. However, advances in large pretrained models make it possible to obtain task representations and descriptions for effective inference and prompt adaptation in image enhancement.

\vspace{.5mm}
\noindent \textbf{Medical imaging applications.}
As a special set of image-level applications, medical imaging frequently encounters distribution shifts, 
e.g., in different hospitals, equipment with different protocols, and even different patients.
Moreover, in clinical practice, the source data usually cannot be accessed during adaptation due to privacy regulations and the label for the target domain is in shortage because of the high cost of professional labeling. 
Therefore, test-time adaptation is becoming more and more common in medical imaging tasks, e.g., classification \cite{ma2022test}, segmentation \cite{bateson2022test, he2021autoencoder, hu2021fully, liu2022single, valanarasu2022fly, varsavsky2020test, wang2023feature}, reconstruction \cite{zhaoe2024test}, registration \cite{zhou2024testa} and accelerated MRI \cite{darestani2022test, klug2024motionttt}.
Most methods rely on model adaptation, often enhanced by specific objective functions tailored for medical data \cite{hu2021fully, darestani2022test}.
Adaptive-UNet \cite{valanarasu2022fly} utilizes inference adaptation at test time by generating the model parameters according to each target sample. 
Since medical imaging applications are similar to common image classification and segmentation tasks, normalization and prompt adaptation methods are also applicable \cite{liu2024zero}. Additionally, applying sample adaptation to medical applications requires sufficient medical data and appropriate tailoring of the method to accurately model the source distributions.

\vspace{.5mm}
\noindent \textbf{Other applications.}
In addition to the previous tasks, test-time adaptation is also applied in other image-level tasks. For instance,  pose estimation \cite{kan2022self, lee2023ttacope, li2021test, kan2023self, cui2023meta}, person re-identification \cite{han2022generalizable, wang2024heterogeneous}, deep fake detection \cite{chen2022ost, zhou2024test}, out-of-distribution detection \cite{gao2023atta, kim2024model, fan2024test}, style transfer \cite{kim2022deep}, and federated learning \cite{feng2023towards, bao2023adaptive, tan2023heterogeneity}.

\vspace{.5mm}
\noindent 
Model adaptation is the most widely utilized approach in current image-level applications. Normalization adaptation is also applicable to models with batch normalization layers. Furthermore, advances in large models have made it easier to obtain more representative task representations and descriptions, benefiting inference adaptation and prompt adaptation methods. Sample adaptation is more suitable for semantic tasks like classification than for dense prediction tasks, which require detailed information. However, with continued advances in generative modeling, sample adaptation may prove a valuable test-time-adaptation approach in the near future, also for tasks beyond image classification.


\subsection{Video-level applications}

Beyond images, video data is gaining increasing attention in computer vision and machine learning. 
Compared to images, video data is more susceptible to distribution shifts due to noise, motion blur, and compression artifacts \cite{lin2022video}. 
Annotating video data is also more challenging, particularly for dense tasks such as segmentation and depth estimation. 
Consequently, adapting to unseen data with unknown distribution shifts is crucial for video applications. 
Despite these challenges, videos inherently contain more information about the data distribution than images, which may results in better adaptation.

\vspace{.5mm}
\noindent \textbf{Action and behavior classification.}
One of the most extensively studied tasks in video applications is action analysis. To address distribution shifts in video data, test-time adaptation methods have been introduced for action and behavior analysis tasks, including classification tasks such as action recognition \cite{lin2022video, xiong2024modality, yi2022temporal} and facial expression recognition \cite{mutlu2023tempt}, as well as temporal action localization \cite{liberatori2024test}. 
Most of these methods involve model adaptation techniques with entropy minimization or pseudo-labeling. Unlike image-level tasks, these methods incorporate video-specific techniques such as temporal consistency \cite{mutlu2023tempt, yi2022temporal} and temporal augmentation \cite{lin2022video}.

\vspace{.5mm}
\noindent \textbf{Dense prediction tasks.}
Beyond classification tasks like action recognition, test-time adaptation is also leveraged in video-level dense prediction tasks, such as video segmentation \cite{azimi2022self, zhang2022auxadapt, bertrand2023test, liu2024depth}, detection \cite{an2024context}, depth prediction \cite{liu2023meta}, and multiple object tracking \cite{segu2023darth}.
Most methods are built on model adaptation, with video-specific supervision like cycle consistency loss of different video frame orders \cite{bertrand2023test}. 

\vspace{.5mm}
\noindent \textbf{Video enhancement.}
There are also test-time adaptation methods working on enhancing the video quality \cite{he2024domain, yang2024genuine}. For example, video denoising to reduce the noise in video frames \cite{yang2024genuine} and video frame interpolation to generate intermediate frames between existing video frames to create smoother motion or higher frame rates \cite{cho2024tta}. 

\vspace{.5mm}
\noindent
Most test-time adaptation methods applied to video tasks are based on model adaptation, as video data provides more information than images, which enhances the fine-tuning processes with self-supervision.  However, beyond model adaptation, the richer (multimodal) information within target data provides more task-specific details, which can also benefit inference, normalization, and prompt adaptation methods. These areas are worth exploring in future research. In contrast, sample adaptation is more challenging to apply in video applications due to the complexity of video data and the difficulty in spatio-temporal generative modeling of the source distribution.

\input{tables/app}

\subsection{3D-level applications}
Beyond the image and video applications, 3D-level applications also face distribution shifts, which require adaptation at test time.
Test-time adaptation for 3D applications currently focuses on tasks like classification and segmentation. 

\vspace{.5mm}
\noindent \textbf{3D classification.} 
3D classification tasks make predictions on a 3D point cloud, which can be corrupted at test time.
The test-time adaptation methods for 3D classification tries to solve distribution shifts in 3D-level through either model adaptation \cite{mirza2023mate, shim2024cloudfixer} or inference adaptation \cite{wang2024backpropagation}.

\vspace{.5mm}
\noindent \textbf{3D dense prediction.} 
In addition to classification, 3D dense segmentation \cite{cao2023multi, cao2024reliable, saltori2022gipso, shin2022mm, weijler2024ttt, prabhudesai2023test} and detection \cite{yuan2024reg, lin2024monotta} are popular applications for test-time adaptation. Methods for 3D segmentation, like those for 2D dense prediction tasks, primarily rely on model adaptation techniques. These include approaches using 3D reconstruction \cite{prabhudesai2023test} or pseudo-labeling \cite{saltori2022gipso, zou2024hgl} at test time. Some methods achieve 3D segmentation with both image and point cloud modalities \cite{shin2022mm, cao2023multi, cao2024reliable, weijler2024ttt}. 

\vspace{.5mm}
\noindent \textbf{Other applications.}
Recently, test-time adaptation approaches have also been investigated in other 3D-level applications.
For 3D pose analysis, test-time adaptation is utilized for pose estimation \cite{zhang2020inference} and human pose forecasting \cite{cui2023test, cui2024human}.
In addition, methods for point-cloud registration \cite{hatem2023point}, flow estimation \cite{zhang2024dual2}, human mesh reconstruction \cite{nam2023cyclic, nie2024incorporating}, and multi-task point cloud understanding \cite{jiang2024pcotta} have also been studied. 

\vspace{.5mm}
\noindent
Most methods applied to 3D-level applications are again based on model adaptation at test time. However, similar to video-level applications, inference and normalization adaptation methods are also applicable to 3D applications with richer information in 3D data. 
Additionally, sample adaptation may require further techniques for modeling 3D inputs or features during source training. To implement prompt adaptation, it is also necessary to design 3D-level prompts or align 3D features to other modalities. Overall, beyond model adaptation, other types of adaptation methods are also applicable to 3D applications and merit exploration in future research.

\begin{tikzpicture}[remember picture, overlay]
\node[fill=yellow!50, opacity=0.3, text opacity=0.5, rounded corners=5pt, minimum width=10.6cm, minimum height=3.8cm] (box) at (10.8,12) {};
\node[text opacity=0.5, anchor=center] at (box.center) [yshift=-5pt, align=center] 
{Emerging research opportunity II: \\ \textbf{Beyond model adaptation and image classification}};
\end{tikzpicture}

\subsection{Beyond vision applications}
Beyond vision applications, test-time adaptation is also appearing more and more in other tasks, e.g., reinforcement learning \cite{hansen2020self, liu2023design, son2023meta, yang2024movie},  natural language processing \cite{banerjee2021self,tula2023or, ye2023robust}, and multimodal learning \cite{huang2023improving, wen2023test, yang2024test}.

\vspace{.5mm}
\noindent
\textbf{Reinforcement learning.} 
Reinforcement learning is a type of machine learning where an agent learns to make decisions by taking actions in an environment. 
Since the agents are typically deployed in dynamic environments that are not covered during training \cite{hansen2020self}, it is crucial to enable the agent to learn and adapt to these new environments on the fly, improving robustness and generalization  \cite{wang2024robust, wang2024simple}.
Therefore, test-time adaptation methods are investigated in reinforcement learning tasks.
Test-time adaptation for reinforcement learning mainly focuses on policy adaptation \cite{hansen2020self, liu2023design, peng2023diagnosis, yang2024movie} and combinatorial optimization \cite{son2023meta}, where model adaptation approaches are usually applied.

\vspace{.5mm}
\noindent
\textbf{Natural language processing.} 
Language data also faces distribution shifts, such as texts from different sources and specific fields, language over time and culture, unique contents of individual users, and texts with corruption.
Therefore, Natural Language Processing (NLP) applications also require test-time adaptation to deal with distribution shifts during deployment.
In NLP applications, test-time adaptation has been utilized in tasks like question answering \cite{banerjee2021self, ye2023robust}, Text-to-SQL \cite{varma2023conditional}, and large language model adaptation \cite{hardt2024test}.
Most of the methods are based on model adaptation.
Since the NLP tasks are different from the visual tasks, unique techniques of the NLP tasks are utilized for adaptation, such as synthetic question generation \cite{banerjee2021self}, related neighbor retrieval \cite{hardt2024test}, and conditional tree matching \cite{varma2023conditional}. 

\vspace{.5mm}
\noindent
\textbf{Multimodal learning applications.}
A large part of recent multimodal learning applications of test-time adaptation focuses on vision-language models for robust visual tasks, adapting prompts at test time \cite{shu2022test, samadh2023align, ma2024swapprompt, fahes2023poda, zhang2022tempera, xiao2024any, zanella2024test}.
In addition to these methods, test-time adaptation is also utilized in visual document understanding \cite{ebrahimi2022test, tula2023or}, vision question answering \cite{wen2023test, liu2024question}, vision-and-language navigation \cite{gao2024fast}, based on model or sample adaptation.

\vspace{.5mm}
\noindent
\textbf{Other tasks.}
In addition to the previous tasks, test-time adaptation has also been introduced into applications like speech \cite{kim2023sgem, kim2022variational, dumpala2023test}, forecasting \cite{park2024t4p, arik2022self, chen2024test, cui2023test}, tabular data \cite{ren2024tablog}, security \cite{guan2024backdoor},  etc. 


\vspace{.5mm}
\noindent
In conclusion, since distribution shifts are prevalent in real-world applications, test-time adaptation has the potential to be applied across various tasks with different data types and objectives. Currently, test-time adaptation methods for applications beyond image classification primarily focus on model adaptation, as it can be easily extended to any task by specifically modifying the unsupervised objective functions.
Normalization adaptation is more prevalent in visual tasks since it relies on batch normalization layers in the network. Sample adaptation faces challenges due to the varying data types across different tasks. Feature-level adjustments and advancements in generative foundation models could potentially address the issue in future work. 
Moreover, along with the advances of large pretrained models, achieving efficient test-time adaptation is crucial, where inference adaptation and prompt adaptation merit more attention in future work. Finally, large pretrained models can also provide more representative task descriptions, benefiting inference and prompt adaptation at test time.


%% file: tables/classification_setting.tex
\begin{table*}[t]
\centering
\vspace{-2mm}
\caption{\textbf{Evaluating test-time distribution shifts.} This table summarizes representative test-time adaptation methods according to what they adapt (Section~\ref{method:w}) and the distribution shift they address in their image classification evaluation (Section~\ref{app:classification}), revealing the emerging research opportunity of test-time adaptation beyond model adaptation and covariate shifts (Section~\ref{futurework}).
}
\label{tab: classapp}
	\resizebox{2.\columnwidth}{!}{%
		\setlength\tabcolsep{4pt} 
\begin{tabular}{lccccc}
\toprule
        & \multicolumn{2}{c}{Covariate shifts} & \multirow{2}*{Label shifts} & \multirow{2}*{Conditional shifts} & \multirow{2}*{Joint shifts} \\
        \cmidrule(lr){2-3} 
        & Single-source & Multi-source & & & \\ 
        \midrule
        \rowcolor{mColor2}
        \multicolumn{6}{l}{\textbf{Model adaptation}} \\
        Auxiliary self-supervision & \cite{sun2020test, liu2021ttt++, gandelsmantest2022test, bartler2022mt3} & \cite{chen2023improved} & - & - & - \\ 
        Entropy minimization & \cite{wang2021tent, zhang2021memo, choi2022improving} & \cite{choi2022improving, zhang2023domainadaptor} & \cite{lee2022surgical, park2023label, zhou2023ods} & & \cite{lee2022surgical, park2023label, gao2024unified, lee2023towards}  \\
        Pseudo-labeling & \cite{goyal2022test, tomar2023tesla, chen2022contrastive} & \cite{jang2022test, ambekarvariational, wang2023feature} & - & - & - \\
        Feature alignment & \cite{jung2022cafa, su2022revisiting, fleuret2021test, nguyen2023tipi} & \cite{wang2023feature} & - & - & - \\
        \midrule
        \rowcolor{mColor2}
        \multicolumn{6}{l}{\textbf{Inference adaptation}} \\
        Batch-wise inference & \cite{zhang2023adanpc, huangextrapolative} & \cite{iwasawa2021test, dubey2021adaptive, zhang2023adanpc} & - & - & - \\
        Sample-wise inference & - & \cite{xiao2022learning, sun2022dynamic} & - & - & - \\ 
        \midrule
        \rowcolor{mColor2}
        \multicolumn{6}{l}{\textbf{Normalization adaptation}}  \\
        Target statistics & \cite{gong2022note, zhang2021adaptive, nado2020evaluating} & - & - & - & - \\
        Statistics combination & \cite{schneider2020improving, lim2023ttn} & - & - & - & - \\
        Statistics inference & - & \cite{du2020metanorm, jiang2023domain} & - & - & -  \\ 
        \midrule
        \rowcolor{mColor2}
        \multicolumn{6}{l}{\textbf{Sample adaptation}} \\
        Feature adjustment & - & \cite{pandey2021generalization, xiao2023energy, park2023test} & - & - & - \\ 
        Input adjustment & \cite{gao2022back} & - & - & - & - \\ 
        \midrule
        \rowcolor{mColor2}
        \multicolumn{6}{l}{\textbf{Prompt adaptation}} \\
        Text-prompting & \cite{hou2024domainverse} & - & \cite{menon2022visual, maniparambil2023enhancing, hou2024domainverse} & - & \cite{hou2024domainverse} \\
        Embedding-prompting & \cite{shu2022test, samadh2023align, gao2022visual, xiao2024any} & \cite{xiao2024any} & \cite{xiao2024any} & \cite{xiao2024any} & \cite{xiao2024any} \\
\bottomrule
\end{tabular}
}
\end{table*}

%% file: tables/app.tex
\begin{table*}[t]
\centering
\vspace{-2mm}
\caption{\textbf{Test-time adaptation applications.} This table summarizes representative test-time adaptation methods according to what they adapt (Section~\ref{method:w}) and their main applications (Section~\ref{application}) revealing the emerging research opportunity of test-time adaptation beyond model adaptation and image classification (Section~\ref{futurework}). 
}
\label{tab: app}
	\resizebox{2.\columnwidth}{!}{%
		\setlength\tabcolsep{4pt} 
\begin{tabular}{lccccccccccc}
\toprule
         & \multicolumn{3}{c}{Image-level} & \multicolumn{3}{c}{Video-level}  & \multicolumn{2}{c}{3D-level} &  \multicolumn{3}{c}{Beyond-vision} \\
        \cmidrule(lr){2-4} \cmidrule(lr){5-7} \cmidrule(lr){8-9} \cmidrule(lr){10-12}
        & Classification & Dense & Enhance & Classification & Dense & Enhance & Classification & Dense & RL & NLP & Multimodal \\ 
        \midrule
        \rowcolor{mColor2}
        \multicolumn{7}{l}{\textbf{Model adaptation}} & & & & & \\
        Auxiliary self-supervision  & \cite{sun2020test, liu2021ttt++} & \cite{park2024test} & \cite{chi2021test, roy2023test} & - & \cite{bertrand2023test, liu2024depth} & \cite{cho2024tta, yang2024genuine} & \cite{mirza2023mate} & \cite{weijler2024ttt,prabhudesai2023test} & \cite{liu2023design, son2023meta} & \cite{hardt2024test, banerjee2021self} & \\ 
        Entropy minimization & \cite{wang2021tent, zhang2021memo} & \cite{sinha2023test, lee2023towards} & - & \cite{yi2022temporal} & - & - & - & -  & - & - & \cite{wen2023test, liu2024question}  \\
        Pseudo-labeling & \cite{jang2022test, goyal2022test} & \cite{veksler2023test, izquierdo2023sfm} & - & \cite{xiong2024modality, liberatori2024test} & - & - & - & \cite{saltori2022gipso, shin2022mm} & - & - & \cite{ebrahimi2022test, tula2023or}  \\
        Feature alignment & \cite{su2022revisiting, mirza2022actmad} & \cite{hong2021deep} & \cite{deng2023efficient} & \cite{mutlu2023tempt, lin2022video} & \cite{azimi2022self, segu2023darth} & - & - & - & - & - & - \\
        \midrule
        \rowcolor{mColor2}
        \multicolumn{7}{l}{\textbf{Inference adaptation}} & & & & &  \\
        Batch-wise inference & \cite{iwasawa2021test, zhang2023adanpc} & - & - & - & - & - & \cite{wang2024backpropagation} & - & - & \cite{varma2023conditional} & - \\
        Sample-wise inference & \cite{xiao2022learning, sun2022dynamic} & \cite{wang2023dynamically, yeo2023rapid} & - & - & - & - & - & - & - & - & - \\ 
        \midrule
        \rowcolor{mColor2}
        \multicolumn{7}{l}{\textbf{Normalization adaptation}} & & & & &  \\
        Target statistics & \cite{gong2022note, zhang2021adaptive} &  - &  - & - & - & - & - & - & - & - & -  \\
        Statistics combination & \cite{schneider2020improving, lim2023ttn} & \cite{khurana2021sita, zou2022learning} & \cite{yao2023generalized} & - &  - & - & - & - & - & - &  - \\
        Statistics inference & \cite{du2020metanorm, jiang2023domain} & - & - &  - & - & - & - & - & - & - &  - \\ 
        \midrule
        \rowcolor{mColor2}
        \multicolumn{7}{l}{\textbf{Sample adaptation}} & & & & & \\
        Feature adjustment & \cite{pandey2021generalization, xiao2023energy} & - & - & - & - & - & - & - & - & - & \cite{agarwal2023star}  \\ 
        Input adjustment & \cite{gao2022back} & - & - & - & - & - & - & - & - & - & - \\ 
        \midrule
        \rowcolor{mColor2}
        \multicolumn{7}{l}{\textbf{Prompt adaptation}} & & & &  & \\
        Text-prompting & \cite{ren2024chatgpt, mirza2024meta} & \cite{fahes2023poda} & - & - & - & -& -& -& -& -& \cite{manas2024improving, zhang2022tempera} \\
        Embedding-prompting & \cite{sun2023vpa, tsai2024convolutional} & - & \cite{chen2024prompt} & - & - & -& -& -& -& -& \cite{shu2022test, samadh2023align} \\
\bottomrule
\end{tabular}
}
\end{table*}

%% file: 6_discussion.tex
\section{Emerging Research Opportunities}
\label{futurework}

From the analysis of current test-time adaptation methods and the distribution shifts they address in Section \ref{app:classification} and their applications in Section \ref{application}, we highlight two emerging research opportunities in this section, divided into test-time adaptation beyond model-adaptation and covariate shifts and test-time adaptation beyond model-adaptation and image classification.

\subsection{Beyond model adaptation and covariate shifts}

\vspace{.5mm}
\noindent \textbf{Distribution shift mixtures.}
Most current test-time adaptation methods focus on covariate shifts, where differences between training and test distributions arise in the input space. However, real-world applications often encounter various distribution shifts between training and test data. For instance, label shifts make the label distribution in the test data differ from the training data, e.g., long-tailed training data. Methods that can incorporate new information without losing previously learned knowledge are crucial for handling label shifts.
Conditional shifts occur when the relationships between features and labels differ across training and test distributions, e.g., spurious correlations or distinct subpopulations, where methods are required to learn conditional dependencies between inputs and outputs at test time.
Moreover, in practice, different types of distribution shifts exist either individually or jointly. Current methods designed for specific shifts can struggle in complex application scenarios, making it essential to explore approaches that can adaptively handle multiple distribution shifts in the future.

\vspace{.5mm}
\noindent \textbf{Open-set.}
Beyond common covariate shifts and label shifts, the test distributions can even involve unseen labels that were not present during training. This scenario becomes more challenging with unseen annotations, as both the input and label spaces differ entirely from the training data, creating an open-set setting that is much more complex than the closed-set case that is currently common.
With recent advances in foundation models \cite{radford2021learning} and their capabilities in zero-shot learning, it is becoming feasible to handle open-set adaptation by leveraging pretrained knowledge within the model.

\vspace{.5mm}
\noindent \textbf{Theoretical analysis.}
Current test-time adaptation methods have demonstrated effectiveness in managing distribution shifts. However, most approaches are empirical and emphasize technical innovations. They often lack a deeper theoretical foundation. For instance, recent studies find empirical evidence that parameters of different layers affect various types of distribution shifts differently \cite{lee2022surgical}, but a further theoretical analysis is needed to guide fundamental understanding. With a solid theoretical understanding of different distribution shifts, it may be possible to derive proven solutions for specific shifts and even develop adaptive methods capable of jointly addressing multiple shifts.

\subsection{Beyond model adaptation and image classification}

\vspace{.5mm}
\noindent \textbf{Foundation models.} 
To tackle various complex tasks beyond image classification, it is crucial to develop more general and powerful models. Recent advancements have made significant progress with various large-scale foundation models \cite{achiam2023gpt, radford2021learning, kirillov2023segment}.
Trained with vast amounts of parameters on extensive datasets, these methods introduce both new research opportunities and challenges for test-time adaptation.
Test-time adaptation enables the foundation models to adapt to specific downstream tasks without specific annotations. However, the approaches must be specifically designed for the large amount of parameters, computational costs, and data heterogeneity.
Model adaptation methods are not ideally suited for foundation models due to their high parameter counts, necessitating careful selection of parameter subsets for adaptation or the use of parameter-efficient fine-tuning techniques like LoRA \cite{hu2021lora}. Normalization adaptation also faces challenges, as large models often replace batch normalization with alternative normalization techniques. In contrast, inference, sample, and prompt adaptation methods may be more compatible with foundation models, though these approaches are still barely being explored.

\vspace{.5mm}
\noindent \textbf{Multi-modal and multi-task scenarios.} 
Beyond applying test-time adaptation to individual applications, there are also practical multi-modal or multi-task scenarios. 
In these cases, test-time adaptation for each modality or task can benefit from the shared information through knowledge transfer or representation alignment, enhancing overall adaptation performance. 
Similar to foundation models, model adaptation and normalization adaptation can be further tailored to multi-modal or multi-task applications. Inference, sample, and prompt adaptation also present potential opportunities for future research.

\vspace{.5mm}
\noindent \textbf{Efficiency and robustness.}
For test-time adaptation of multimodal foundation models, achieving efficient and robust adaptation at test time is a must. 
Due to the large amount of parameters, adjusting the models requires high computational costs and vast amounts of data, which are often impractical at test time. Future research could investigate lightweight adaptation strategies, such as parameter-efficient fine-tuning (e.g., adapters, low-rank matrices) or selective adaptation, where only the most critical parameters are adjusted. This enables efficient adaptation without excessive computational overhead, making adaptation viable for resource-constrained application scenarios like robotics. Additionally, limited data at test time can lead to overfitting or overconfident predictions. Therefore, it is also essential for future research to avoid overconfidence and overfitting at test time, while maintaining the original generalization ability of the pretrained model. Finally, utilizing pretrained multimodal foundation models as assistants that benefit the adaptation and inference at test time is also a promising research direction.

%% file: 7_conclusion.tex
\section{Conclusion}
\label{conclusion}
Adaptation across distributions at test time has wide practical applications and is becoming an emerging and popular topic in machine learning.
In this paper, we provide a comprehensive survey of existing test-time adaptation methods, categorizing them into model, inference, normalization, sample, and prompt adaptation approaches. We discuss each adaptation type to highlight its strengths and challenges, as well as common training preparation and adaptation settings of various approaches. We also summarize the evaluation and applications of current test-time adaptation methods, according to which we identify emerging research opportunities, especially for approaches beyond model adaptation. 
